\documentclass[11pt]{article}

\usepackage[preprint]{acl}

\usepackage{times}
\usepackage{latexsym}
\usepackage[T1]{fontenc}
\usepackage[utf8]{inputenc}
\usepackage{microtype}
\usepackage{inconsolata}

\usepackage{amsmath,amssymb}
\usepackage{amsmath,amsfonts,bm}
\usepackage{mathtools}

\def\eqref#1{equation~\ref{#1}}
\def\1{\bm{1}}

\DeclareMathAlphabet{\mathsfit}{\encodingdefault}{\sfdefault}{m}{sl}
\SetMathAlphabet{\mathsfit}{bold}{\encodingdefault}{\sfdefault}{bx}{n}

\usepackage{graphicx}
\usepackage{subcaption}
\usepackage{booktabs}
\usepackage{array}
\usepackage{enumitem}
\usepackage{multirow}
\usepackage{wrapfig}
\usepackage{algorithm}
\usepackage{algpseudocode}
\algrenewcommand{\algorithmicreturn}{\textbf{return }}
\usepackage{fontawesome5}
\usepackage{xcolor}

\newcolumntype{C}[1]{>{\centering\arraybackslash}p{#1}}
\newcolumntype{P}[1]{>{\centering\arraybackslash}p{#1}}

\newcommand{\hlr}[1]{#1}
\newcommand{\hly}[1]{#1}
\newcommand{\hlb}[1]{\textbf{#1}}
\newcommand{\dianbo}[1]{}
\newcommand{\alex}[1]{}

\usepackage{fancyhdr}
\fancypagestyle{firstpage}{%
  \fancyhf{}%
  \fancyfoot[L]{%
    \begin{tabular}[b]{@{}l@{}}
      \small$^{*}$Equal contribution.\enspace
        \href{mailto:yks23@tsinghua.edu.cn}{\small\texttt{yks23@tsinghua.edu.cn}},\;
        \href{mailto:zhangtinghe5@gmail.com}{\small\texttt{zhangtinghe5@gmail.com}}\\
      \small$^{\dagger}$Corresponding author.\enspace
        \href{mailto:lambalex@tsinghua.edu.cn}{\small\texttt{lambalex@tsinghua.edu.cn}}
    \end{tabular}%
  }%
  \fancyfoot[C]{\small\thepage}%
}
\fancypagestyle{ttspage}{%
  \fancyhf{}%
  \fancyfoot[L]{%
    \parbox[b]{0.9\textwidth}{\small$^{\ddagger}$An earlier version of this paper reported incorrect test-time scaling results due to an experimental error. all results in this version are re-verified.}%
  }%
  \fancyfoot[R]{\small\thepage}%
}

\begin{document}

\twocolumn[{%
\vskip 0.15in
\begin{center}
{\Large\bfseries Explore-Execute Chain: Towards an Efficient Structured Reasoning Paradigm}\\[1.2em]

\textbf{Kaisen Yang}$^{*,1,2,3}$ \enspace
\textbf{Tinghe Zhang}$^{*,1,3,7}$ \enspace
\textbf{Rushi Shah}$^{4}$ \enspace
\textbf{Kaicheng Yang}$^{5}$ \enspace
\textbf{Qinwei Ma}$^{1,3,6}$ \enspace \\
\textbf{Dianbo Liu}$^{4}$ \enspace
\textbf{Alex Lamb}$^{1,3\dagger}$
\\[0.7em]

{\small
$^{1}$Qizhi Institute\enspace
$^{2}$Dept.\ of Computer Science, Tsinghua University\enspace
$^{3}$College of AI, Tsinghua University\enspace
$^{4}$Engineering Department, National University of Singapore\enspace
$^{5}$Dept.\ of Automation, Shanghai Jiao Tong University\enspace
$^{6}$IIIS, Tsinghua University\enspace
$^{7}$College of Software, Northeastern University
}\\[0.8em]
% Clickable resource links
{\small
\href{https://github.com/OliverZ-dot/Explore-Execute-Chain}%
  {\faGithub~\texttt{Code}}%
\quad\textbar\quad
\href{https://huggingface.co/TingheOliver/Explore-Execute-Chain-Qwen}%
  {\raisebox{-1.5pt}{\includegraphics[height=0.85em]{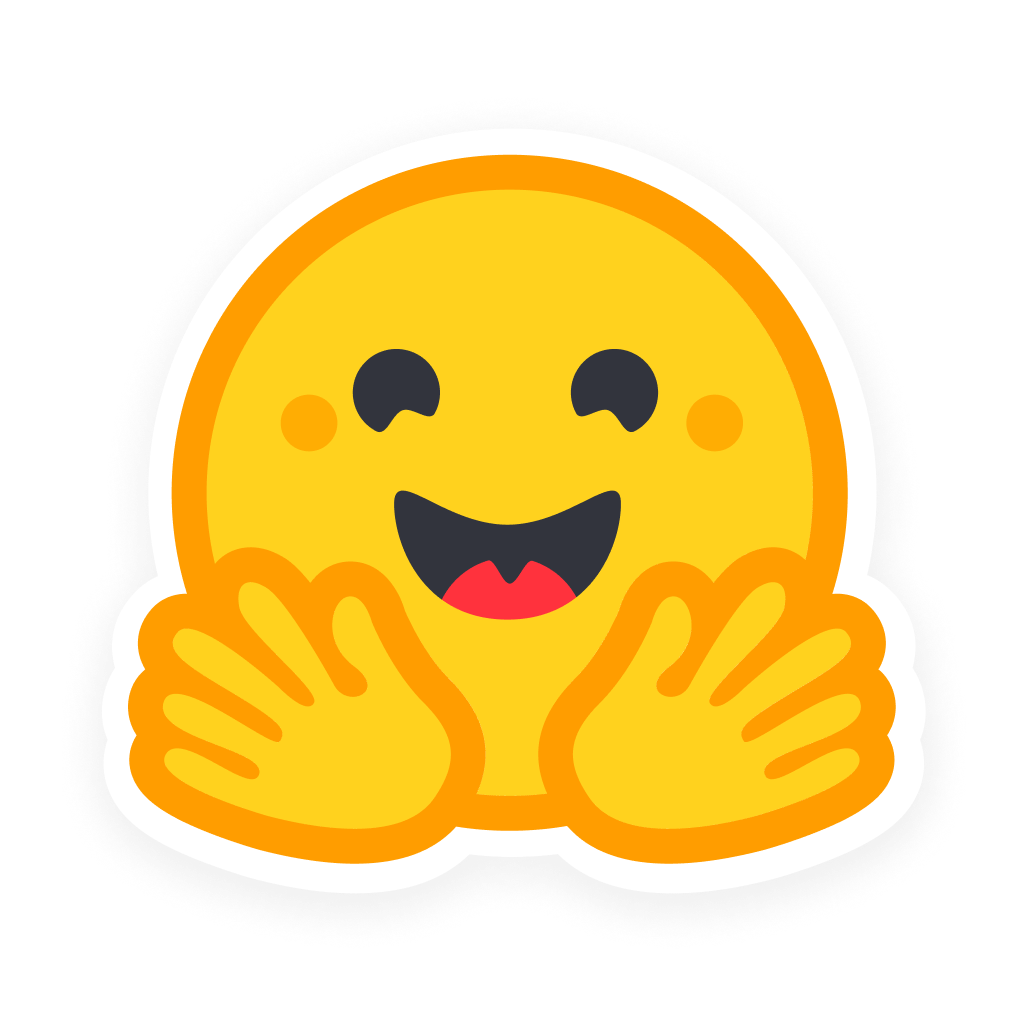}}~\texttt{Model}}%
\quad\textbar\quad
\href{https://huggingface.co/datasets/TingheOliver/Explore-Execute-Chain-Datasets}%
  {\raisebox{-1.5pt}{\includegraphics[height=0.85em]{hf-logo}}~\texttt{Dataset}}%
}
\end{center}
\vskip 0.4in
}]
\thispagestyle{firstpage}

\begin{abstract}
Many LLMs plan before they act, yet planning and execution are often still entangled in one long generation trace, enforced only through prompts, or split across separate components. We argue that these two stages call for different computation: planning benefits from diversity and breadth, whereas execution demands precision and faithful adherence to a chosen strategy. Treating them as a single undifferentiated chain wastes tokens on routine derivation and makes it costly to explore alternative strategies at test time. We present the \textbf{Explore-Execute Chain (E\textsuperscript{2}C)}, which keeps both stages in one model but separates them structurally: a stochastic \textit{Exploration} phase drafts a concise high-level plan, and a deterministic \textit{Execution} phase carries it out. Causal SFT and RL train this split so that exploration stays informative and execution remains plan-faithful. Once plans are short yet decisive, extra inference compute can be directed to exploration rather than to repeatedly decoding full solutions. On AIME'2024 at $K{=}32$, \textbf{E\textsuperscript{2}C-ReAct Loop} reaches 53.3\% accuracy with only 12.4k tokens, outperforming Tree-of-Thoughts ($N{=}32$: 50.0\%, 71.3k). The same structure also supports lightweight domain adaptation: \textbf{Exploration-Focused SFT (EF-SFT)} updates only the planning phase, uses 3.5\% of the tokens required by standard SFT, and improves medical benchmark accuracy by up to 14.5\%.
\end{abstract}

\section{Introduction}

Large Language Models (LLMs) have demonstrated remarkable capabilities in complex reasoning, largely propelled by techniques such as Chain-of-Thought (CoT) prompting~\citep{wei2022chain}. This paradigm has inspired a suite of advanced methods, including sampling multiple reasoning paths for consensus via Self-Consistency~\citep{wang2022self}, and exploring the solution space with more complex structures like Tree-of-Thoughts (ToT)~\citep{yao2023tree}, Graph-of-Thoughts (GoT)~\citep{besta2023graph}, and Forest-of-Thought (FoT)~\citep{bi2025forest}. Other approaches focus on iterative refinement through self-correction~\citep{shinn2023reflexion} or problem decomposition~\citep{zhou2022least, yao2022react}.

Despite their success, most of these methods still generate planning and execution within a single undifferentiated reasoning chain. Prior work has explored plan-then-execute patterns through subtask decomposition~\citep{zhou2022least}, interleaved reasoning and tool use~\citep{yao2022react}, or LLM planners paired with external solvers~\citep{hao2023reasoning, liu2023llm}. However, such approaches typically interleave planning with action at similar granularity, rely on prompting rather than training the decomposition, or delegate execution outside the model. They therefore still conflate two fundamentally different computational needs within reasoning: high-level strategic \textbf{planning}, which benefits from diversity, and low-level step-by-step \textbf{execution}, which requires precision. This entanglement leads to critical inefficiencies. First, the model expends equivalent computational effort on both creative planning and routine calculations~\citep{xu2025adaptive, li2025compressing}. Second, greedy decoding restricts the diversity of initial strategies, where a suboptimal early choice can derail the entire reasoning path~\citep{liao2025fractured, xu2025softcotpp, zheng2025fr3e}.

\begin{figure*}[t]
    \centering
    \begin{minipage}[c]{0.37\textwidth}
        \centering
        \includegraphics[width=\linewidth]{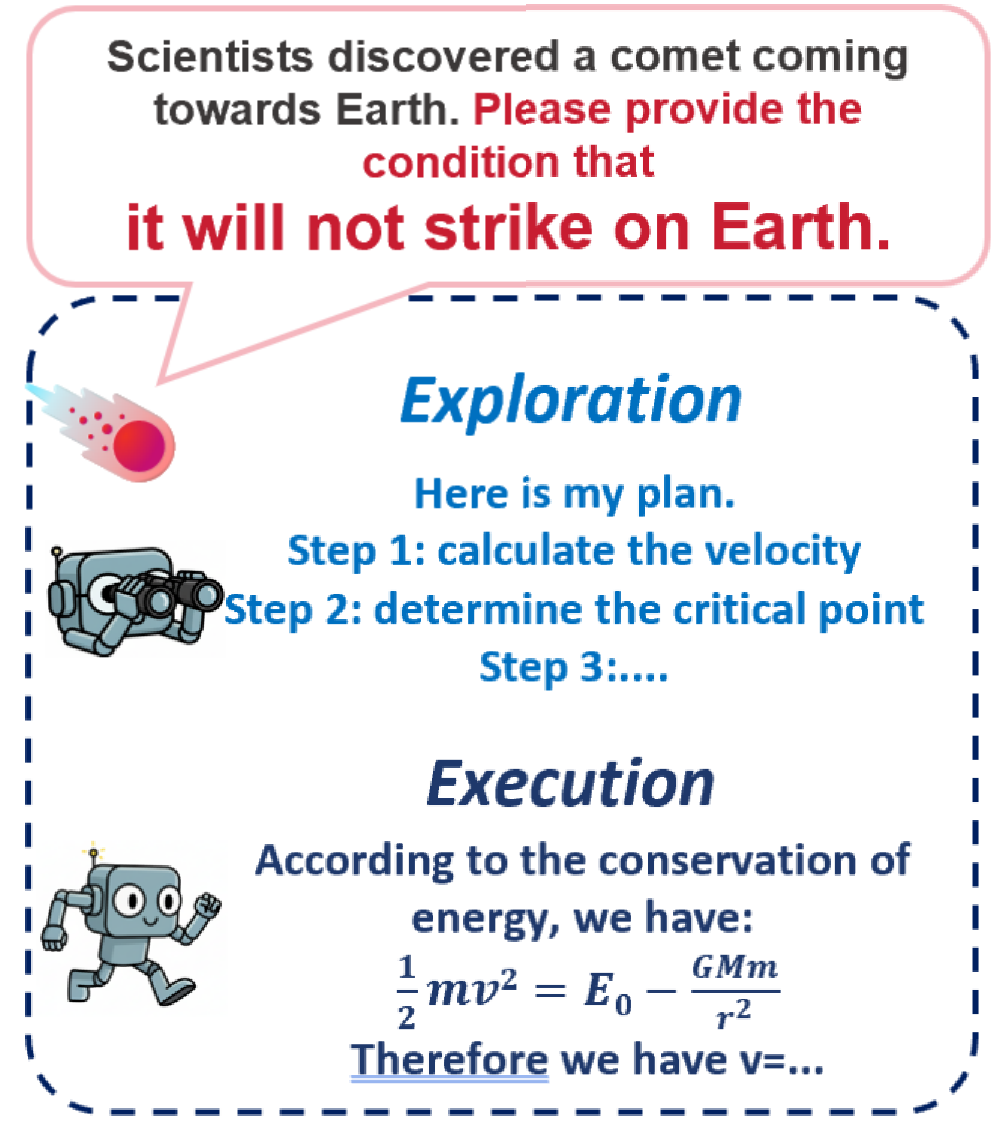}
    \end{minipage}
    \hfill
    \begin{minipage}[c]{0.60\textwidth}
        \centering
        \includegraphics[width=\linewidth]{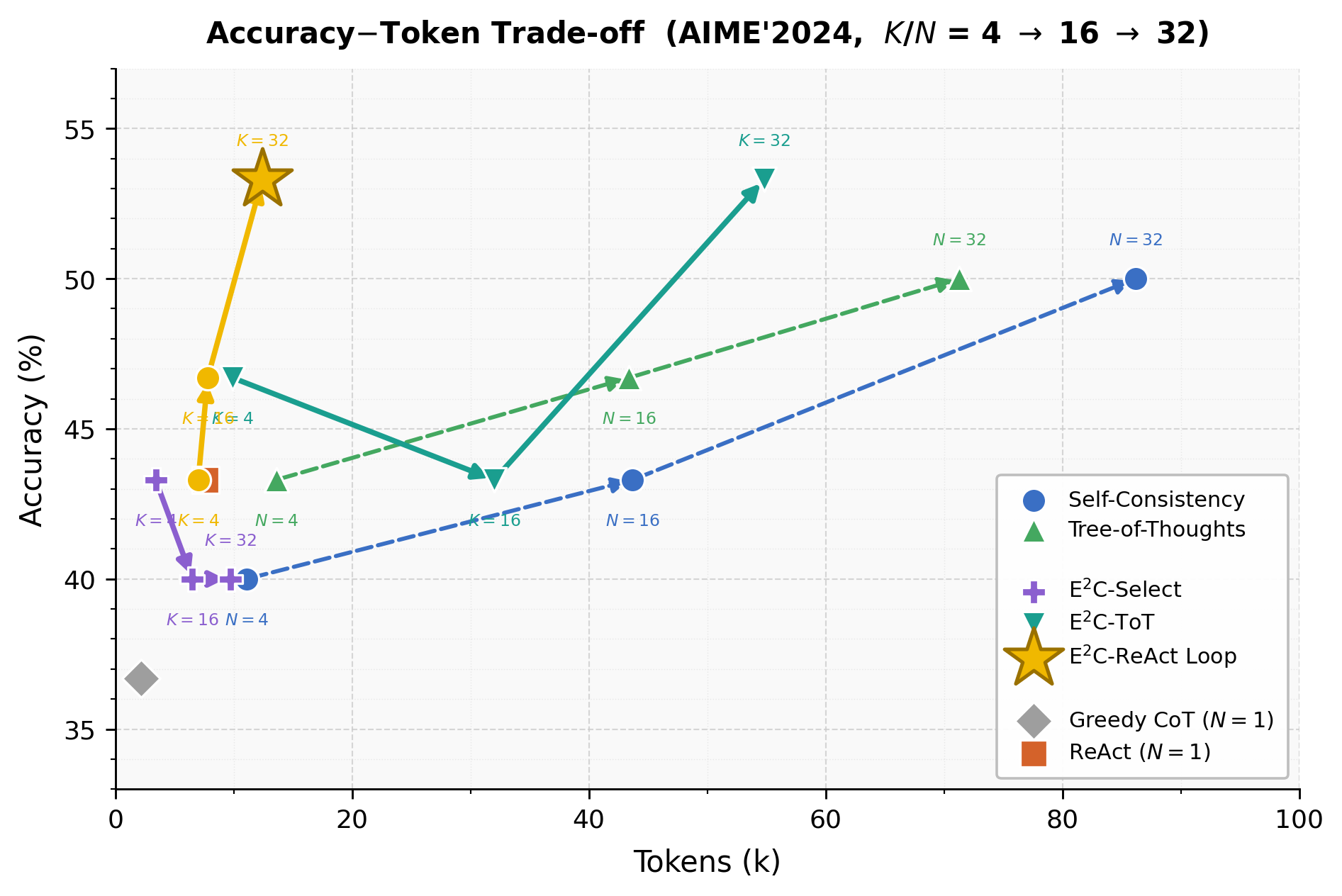}
    \end{minipage}
    \caption{\textbf{Left:} A concrete example of \textbf{E\textsuperscript{2}C} in action. Given a physics question, the model first produces a concise \textit{Exploration} that outlines the high-level solution strategy (e.g., ``Step 1: calculate the velocity; Step 2: determine the critical point...''), then performs a detailed, step-by-step \textit{Execution} (deriving the equations and arriving at the answer). The two stages are structurally separated, keeping planning succinct and computation faithful. \textbf{Right:} Accuracy--token trade-off on AIME'2024 at budget $K$/$N{=}32$. E\textsuperscript{2}C-ReAct Loop (\textbf{\large$\star$}) achieves 53.3\% accuracy with only 12.4k tokens, outperforming all full-chain baselines that require 4--7$\times$ more tokens to reach 50.0\%.}
    \label{fig:teaser}
\end{figure*}
Rather than introducing planning, we ask how a single LLM should internalize plan-then-execute reasoning as a trainable, structurally explicit process. We introduce the \textbf{Explore–Execute Chain (E\textsuperscript{2}C)}, which decomposes standard CoT into two distinct output phases within one model (Figure~\ref{fig:teaser}, left). The first phase is a highly informative \textbf{exploration} stage that produces a concise high-level plan, offering a compact preview of the solution strategy without the cost of full-chain generation. The second phase is a highly deterministic \textbf{execution} stage that carries out the plan in detailed calculations. Unlike prompt-only decomposition, both phases are produced by the same model and are explicitly trained to satisfy complementary objectives: informative planning and faithful execution.

This decomposition enables a highly efficient \textbf{test-time scaling} strategy. Existing methods such as Self-Consistency~\citep{wang2022self}, Tree-of-Thoughts~\citep{yao2023tree}, and ReAct~\citep{yao2022react} all scale full reasoning chains, so their token cost grows in proportion to the budget. E\textsuperscript{2}C breaks this coupling: because Exploration is short and highly informative, additional compute can be concentrated there while Execution remains sparse and deterministic. Each standard TTS strategy has a corresponding E\textsuperscript{2}C variant that operates over plans rather than full chains: multi-sample aggregation becomes plan selection, tree search is restricted to the Exploration stage, and iterative refinement operates over plans rather than complete solutions. As shown in Figure~\ref{fig:teaser} (right), this yields dramatically better accuracy--token efficiency: E\textsuperscript{2}C-ReAct Loop surpasses full-chain baselines while using a fraction of their token budget. We implement this framework using a two-stage (SFT+RL) training pipeline, guided by recent advances in reasoning alignment~\citep{gan2025cotspace, rafailov2023direct}.

Our main contributions are summarized as follows:
\begin{itemize}[leftmargin=15pt,topsep=3pt,itemsep=3pt]
\item We show that, within a single LLM, reasoning can be more effectively organized as two computationally distinct phases: a planning stage that benefits from stochastic diversity and an execution stage that demands precision and determinism. E\textsuperscript{2}C makes this separation explicit in both output structure and training, rather than relying on prompt-level plan-then-execute heuristics alone.
\item We introduce a robust two-stage training methodology (SFT+RL) together with a specialized data construction algorithm that ensures the model faithfully adheres to its plans, effectively instilling E\textsuperscript{2}C paradigm and achieving superior performance.
\item We demonstrate the efficiency of this framework through two applications: concentrating inference compute on the cheap Exploration stage yields competitive test-time scaling with substantially fewer tokens, while Exploration-Focused SFT (EF-SFT) enables lightweight domain adaptation with significant accuracy gains.
\end{itemize}

\section{Preliminary}
\label{sec:preliminary}

We introduce the notation used throughout Methodology. Given a context $c$ (e.g., a question) and a target answer $a$, the model generates two structured components before producing $a$. The \textbf{Exploration} plan $\pi$ is a concise high-level strategy. The \textbf{Execution} trace $e=(e_1,\dots,e_T)$ is the detailed step-by-step derivation that follows $\pi$, where $T$ denotes the execution length.

Standard Chain-of-Thought (CoT) couples these two roles in a single autoregressive chain modeled by $p(e\mid c)$. E\textsuperscript{2}C instead factorizes reasoning within one LLM as
\begin{equation}
p'(\pi,e\mid c)=p'(\pi\mid c)\,p'(e\mid \pi,c).
\label{eq:prelim_factorization}
\end{equation}
The distribution $p'(\pi\mid c)$ controls plan generation and should remain short yet informative. The distribution $p'(e\mid \pi,c)$ controls execution and should follow the chosen plan faithfully. Section~\ref{sec:formal_e2c} formalizes these requirements and describes how training enforces them.

	\section{Methodology}

\begin{figure*}[t]
\centering
\includegraphics[width=\textwidth]{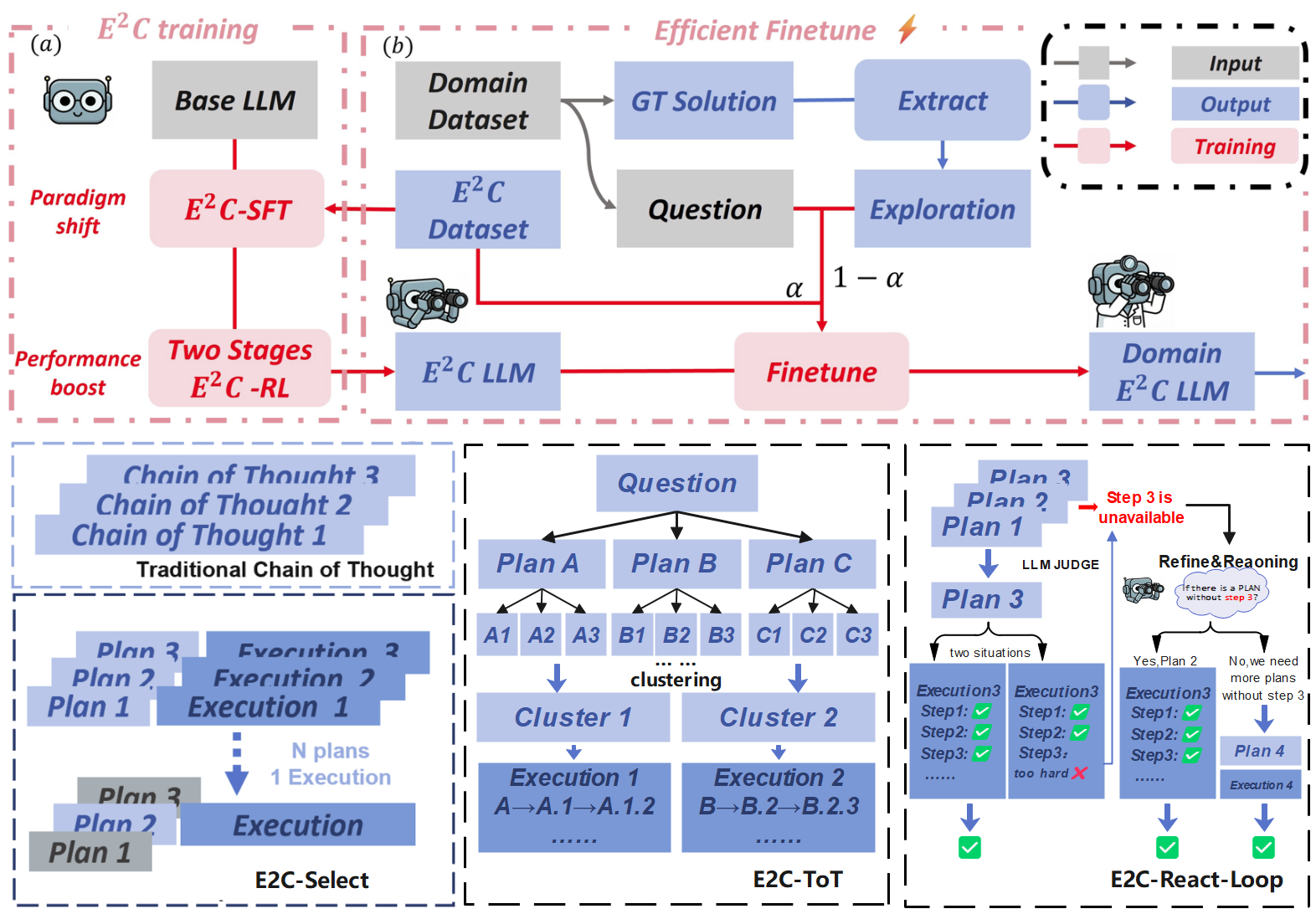}

\caption{\textbf{Overview of E\textsuperscript{2}C method.} The approach begins with E\textsuperscript{2}C-SFT to achieve a paradigm shift, followed by a two-stage E\textsuperscript{2}C-RL process that leverages the decomposition advantage of the new paradigm to boost performance. The resulting E\textsuperscript{2}C-LLM can be efficiently adapted to new domains via EF-SFT. At test time, the high informativeness and compact length of the Exploration stage allows additional compute to be concentrated on plan-level search, selection, and refinement rather than on costly full-chain generation.}

\label{fig:method}

\end{figure*}

	Building on this formulation, we introduce the \textbf{Explore-Execute Chain (E\textsuperscript{2}C)} framework. As shown in Fig.~\ref{fig:method}, we first describe a two-stage training procedure, then present EF-SFT and plan-level test-time scaling.

	\subsection{Formal Definition of E\textsuperscript{2}C}
	\label{sec:formal_e2c}
	
	Building on Eq.~\eqref{eq:prelim_factorization}, E\textsuperscript{2}C replaces the coupled process $p(e\mid c)$ with the explore-execute factorization
	\begin{equation}
\resizebox{\columnwidth}{!}{$\displaystyle
\underbrace{p(e \mid c)}_{\text{Coupled Reasoning Process}} \rightarrow
\underbrace{p'(\pi, e \mid c)}_{\text{Explore-Execute Chain}} =
\underbrace{p'(\pi \mid c)}_{\textcolor{red}{\text{Highly Informative}}} \cdot
\underbrace{p'(e \mid \pi, c)}_{\textcolor{blue}{\text{Highly Deterministic}}}$}
\end{equation}
    The framework is defined by two core properties:
\begin{enumerate}[leftmargin=25pt,topsep=3pt, itemsep=1pt]
\item \label{req:informative} \textbf{(Informative Property)}. $p'(\pi \mid c)$ should be highly informative, containing the critical information necessary to solve the problem.
\item \label{req:deterministic} \textbf{(Deterministic Property)}. $p'(e \mid \pi, c)$ should be highly deterministic, meaning it must fully leverage the informative $\pi$.
\end{enumerate}
Naturally, we semantically design $\pi$ to represent high-level strategies, while $e$ entails detailed calculations that follow $\pi$.
    %\vspace{-1mm}
	\subsection{2-Stage Training Procedure: SFT and RL}
    %\vspace{-2mm}
	We introduce a two-stage training procedure to achieve the proposed Prop.~\ref{req:informative} and Prop.~\ref{req:deterministic}. Stage 1 is Supervised Fine-Tuning (SFT), in which we construct a synthetic dataset and perform SFT to achieve a paradigm shift in reasoning and satisfy the informative Prop.~\ref{req:informative}. We do not rely solely on prompting to accomplish this paradigm transition because prompting is unstable and leads to a more significant performance drop compared to SFT training. Detailed results are presented in Table~\ref{tab:math_performance}.
    Stage 2 employs Reinforcement Learning (RL), which incorporates a $\lambda$-coefficient on the advantage to appropriately leverage Prop.~\ref{req:informative}, thereby accelerating convergence and enhancing the determinism of execution to satisfy Prop.~\ref{req:deterministic}.
  
	%\subsubsection{Stage 1: Synthetic dataset construction and E\textsuperscript{2}C Paradigm-shifting SFT(E\textsuperscript{2}C-SFT)}
%\vspace{-3mm}
\subsubsection{Stage 1: E\textsuperscript{2}C Supervised Finetuning}
\label{sec:dataset}
%\vspace{-2mm}
To support structured reasoning, we construct a dedicated SFT dataset through synthetic generation. A naive method is to first sample an execution trace from the base model and then summarize it into an exploration step. However, this approach is flawed: the execution is generated from $p(e \mid c)$ rather than the desired $p'(e \mid \pi, c)$, effectively hacking the causal structure. As a result, the model learns to ignore the exploration and directly mimic the base model's execution distribution, violating the intended information bottleneck.

Our method, formalized in Algorithm~\ref{alg:correct_sft}, reverses this order: for each question, we first generate a full solution, distill it into an exploration step, and then prompt the model to produce a new execution conditioned on that plan. This enforces a causal dependency from exploration to execution, which is crucial for Prop.~\ref{req:deterministic}.
The solution can also come from the ground truth. To enable a fair comparison and minimize dataset selection constraints while avoiding the introduction of extra variables, we specifically use samples from the Base LLM in our comparison experiments

%\vspace{-3mm}
\subsubsection{Stage 2: E\textsuperscript{2}C Reinforcement Learning} \label{sec:rl}  
%\vspace{-1mm}
To emphasize informative reasoning, we extend hierarchical weighting~\citep{wang2025emergent} by assigning a higher coefficient $\lambda$ to exploration tokens, which accelerates convergence (Prop.~\ref{req:informative}), while the entropy-reduction effect of reinforcement learning supports determinism (Prop.~\ref{req:deterministic}). The training objective is defined as

\begin{equation}
\resizebox{0.92\columnwidth}{!}{$\displaystyle
\mathcal{L}_{\mathrm{clip}}
= \frac{1}{G}\sum_{i,t}
\min\!\Big(
r_{i,t}\,\textcolor{red}{\lambda_{i,t}}\,\hat{A}_{i,t},\;
\operatorname{clip}(r_{i,t},\,1-\varepsilon,\,1+\varepsilon)\,
\textcolor{red}{\lambda_{i,t}}\,\hat{A}_{i,t}
\Big)$}.
\end{equation}
\begin{equation}
\mathcal{J}_{\mathrm{GRPO}}(\theta)
= \mathbb{E}\!\left[\mathcal{L}_{\mathrm{clip}}\right]
-\beta\, D_{\mathrm{KL}}\!\big[\pi_\theta \,\|\, \pi_{\mathrm{ref}}\big].
\end{equation}

where $\hat{A}_{i,t}=(r_{i,t}-\bar{r}_i)/\sigma_i$ and $r_{i,t}=r_{\text{answer}}+r_{\text{format}}$. The reward $r_{\text{answer}}$ measures answer correctness, while $r_{\text{format}}$ consists of a length reward ($r_{\text{length}}$) designed to prevent overly long and repetitive answers and an instruction reward ($r_{\text{instr}}$), quantifies the alignment between exploration and execution, ensuring that exploration trajectories approximate optimal execution strategies.The detailed description for $r_{\text{format}}$ can be found in Appendix~\ref{app:hyper}.

We adopt a two-stage training procedure. In the first stage, a higher temperature $\tau_1$ and larger rollout number $k_1$ are used for one epoch, encouraging broad exploration of the action space and fostering self-correction to mitigate the overly rigid adherence to the exploration plan that results. In the second stage, we reduce the temperature to $\tau_2$ and the rollout number to $k_2$, again for one epoch, and assign the advantage coefficient $\lambda_{i,t}=\lambda_{exp}>1$ for the exploration tokens in the GRPO update. This modification explicitly prioritizes high-level reasoning in the policy gradient, thereby achieving faster and more stable convergence.

The behavior of the trained agent can be formalized by analyzing the modified GRPO objective in Eq.~(2).  
We highlight the following quantified properties:

\textbf{1. Update emphasis.}
Assigning $\lambda_{i,t}=\lambda_{\text{exp}}>1$ for $t\in T_{\text{exp}}$ and $\lambda_{i,t}=1$ for $t\in T_{\text{exe}}$ yields
\begin{equation}
    \frac{\mathbb{E}\!\left[\|g_{i,t}\|^2\mid t\in T_{\text{exp}}\right]}
         {\mathbb{E}\!\left[\|g_{i,t}\|^2\mid t\in T_{\text{exe}}\right]}
    \;\gtrsim\;\lambda_{\text{exp}}^2,
\end{equation}
so exploration tokens receive disproportionately larger gradient updates, strengthening the planning phase. Entropy dynamics in Appendix~\ref{entropy} confirm the effect of $\lambda_{\text{exp}}$.

\textbf{2. Deterministic execution.}
Stage-2 RL (lower temperature, fewer rollouts) drives
\begin{equation}
\resizebox{0.92\columnwidth}{!}{$\displaystyle
    \mathbb{E}_{t\in T_{\text{exe}}}[\Delta_{i,t}] \;\nearrow,
    \qquad
    \mathbb{E}_{t\in T_{\text{exe}}}\!\big[H(\pi_\theta(\cdot\mid q,o_{<t}))\big] \;\searrow,$}
\end{equation}
yielding faithful, low-variance execution.

\textbf{3. Plan sensitivity.}
The expected update sign for exploration tokens satisfies
\begin{equation}
    \mathbb{E}\!\left[\operatorname{sgn}(g_{i,t})\mid t\in T_{\text{exp}}\right]
    \;\propto\;
    \operatorname{sgn}\!\Big(\mathbb{E}[\hat A^{\text{plan}}_{i,t}]\Big),
\end{equation}
so high-quality plans are amplified while poor plans are suppressed.

\subsection{Efficient Adaptation and Inference with E\textsuperscript{2}C}
\label{sec:adaptation_and_inference}
%\vspace{-1mm}

The modularity of our E\textsuperscript{2}C framework enables efficient strategies for both domain adaptation at training time and scaled aggregation at test time.

\textbf{Exploration-Focused SFT (EF-SFT).} For domain adaptation, we introduce EF-SFT. This method leverages the transferable nature of the execution component by exclusively fine-tuning on the exploration segments from domain-specific examples. These segments are mixed with the base E\textsuperscript{2}C dataset at a controlled ratio $\alpha$, allowing the model to efficiently learn new reasoning strategies while maintaining its core capabilities. This targeted approach significantly reduces the data and computational requirements for adaptation. A detailed algorithm can be found in the Appendix~\ref{alg:EF-SFT}.

\textbf{E\textsuperscript{2}C Test-Time Scaling: Scale Exploration, Execute Selectively.}
Standard TTS methods scale full reasoning chains: Self-Consistency samples $N$ complete solutions, Tree-of-Thoughts builds a search tree over intermediate reasoning steps, and ReAct iteratively reissues full solution attempts. Because Execution dominates the token budget of each chain, the cost of these methods grows proportionally with $N$.

E\textsuperscript{2}C breaks this proportionality. Since the Exploration stage is short ($\leq$512 tokens per plan) yet highly informative, we can afford to generate or revise many plans at a small fraction of the cost of one full Execution. This gives each standard TTS strategy a more efficient E\textsuperscript{2}C counterpart:
\begin{itemize}[leftmargin=15pt,topsep=2pt,itemsep=1pt]
  \item \textbf{E\textsuperscript{2}C-Select (Plan Selection):} the plan-level counterpart of multi-sample aggregation. Sample $K$ exploration plans, use the model as a self LM-judge to select the single best plan, and execute it once.
  \item \textbf{E\textsuperscript{2}C-ToT (Exploration-Only Tree Search):} the plan-level counterpart of tree search over full chains. Build a ToT-style tree solely over the Exploration stage, cluster the resulting plan leaves, and execute only representative ones.
  \item \textbf{E\textsuperscript{2}C-ReAct Loop (Execute-Refine Loop):} the plan-level counterpart of iterative full-chain refinement. Execute a selected plan; if execution stalls or fails to produce a valid answer, revise or replace the plan using the failure as feedback, then retry. All iterative search is confined to the compact plan space.
\end{itemize}
The three variants share a common explore-select-execute-refine template; Algorithm~\ref{alg:tts} in the Appendix formalizes this procedure, with each variant instantiating a different selection or refinement operator.

%\vspace{-3mm}
\section{Experiments}
%\vspace{-4mm}
\subsection{Training Protocols}
\label{setup}
We adapt our training codebase from verl~\citep{verl}. The \textbf{E\textsuperscript{2}C-SFT} model was trained for one epoch on a 50k-sample synthetic dataset built from Openr1-math~\citep{openr1math} with the causal pipeline in Algorithm~\ref{alg:correct_sft}, then further refined with our two-stage \textbf{E\textsuperscript{2}C-RL} on DAPO-17K~\citep{yu2025dapo}. A standard \textbf{GRPO} baseline was trained for five epochs on the same data. For domain adaptation, \textbf{EF-SFT} uses a targeted 50k-sample exploration-only subset mixed with 10\% regularization data, compared against standard SFT trained on the full ReasonMed dataset.
\subsection{Experiments Setup}

\begin{table*}[t!]
\centering
\caption{Performance comparison of Qwen3 models (non-thinking mode) on mathematical reasoning benchmarks. All results are reported as Pass@1 accuracy, with an 8-sample average.}
\label{tab:math_performance}
%\vspace{-2mm}
\resizebox{\textwidth}{!}{%
\begin{tabular}{@{}lccccccccc@{}}
\toprule
\textbf{Model} & \textbf{AIME'24} & \textbf{AIME'25} & \textbf{MATH500} & \textbf{Algebra} & \textbf{Minerva} & \textbf{AMC23} & \textbf{Olympiad} & \textbf{Avg Acc} & \textbf{Avg Length} \\ \midrule
\multicolumn{10}{l}{\textit{Qwen3 8B Series}} \\

Qwen3-8B+GRPO (Baseline) & \hlr{36.9} & \hlb{34.4} & \hlb{88.2} & \hly{88.2} & \hly{33.1} & \hly{79.3} & \hly{60.0} & \hly{60.0} & 1429.46 \\

Qwen3-8B+E\textsuperscript{2}C-SFT+GRPO & \hly{37.5} & \hlr{32.5} & \hlr{83.5} & \hlr{86.6} & \hlr{30.8} & \hlr{76.3} & \hlr{56.8} & \hlr{57.7} & 1309.62 \\

\textbf{Qwen3-8B+E\textsuperscript{2}C-(SFT+RL)} & \hlb{40.6} & \hly{33.8} & \hly{87.7} & \hlb{90.9} & \hlb{35.8} & \hlb{80.3} & \hlb{61.3} & \hlb{61.5} & 1476.41 \\
\midrule
\multicolumn{10}{l}{\textit{Qwen3 4B Series}} \\
Qwen3-4B+GRPO (Baseline) & \hlr{28.8} & \hlb{30.6} & \hlr{84.6} & \hly{84.4} & \hly{33.5} & \hly{75.8} & \hly{57.8} & \hly{56.5} & 1263.15 \\

Qwen3-4B+E\textsuperscript{2}C-SFT+GRPO & \hlr{28.8} & \hlr{26.9} & \hly{85.9} & 
\hlr{83.3} & \hlr{33.2} & \hlr{75.7} & \hlr{55.3} & \hlr{55.2} & 1324.18 \\

\textbf{Qwen3-4B+E\textsuperscript{2}C-(SFT+RL)} & \hlb{37.5} & \hly{30.0} & \hlb{86.1} & \hlb{84.8} & \hlb{34.0} & \hlb{78.3} & \hlb{58.4} & \hlb{58.4} & 1456.34 \\

\midrule
\multicolumn{10}{l}{\textit{Ablation Studies}} \\
Qwen3-8B+Prompt (Zero-shot) & \hlr{21.9} & \hlr{18.8} & \hly{76.3} & \hly{80.5} & \hlr{30.9} & \hlr{50.7} & \hly{45.8} & \hlr{46.6} & 1142.38 \\
Qwen3-8B+E\textsuperscript{2}C-SFT & \hlr{23.1} & \hly{21.9} & \hlr{75.8} & \hlr{80.5} & \hly{31.5} & \hly{51.5} & \hlr{43.2} & \hly{46.8} & 1162.89 \\
\bottomrule
\end{tabular}%
}
\end{table*}

\textbf{Mathematical reasoning.}
\textit{Models:} Qwen3-8B and Qwen3-4B in non-thinking mode, comparing E\textsuperscript{2}C-(SFT+RL) with matched GRPO baselines (Sec.~\ref{setup}).
\textit{Benchmarks:} AIME'24, AIME'25, MATH500, Algebra, Minerva, AMC23, and Olympiad Bench~\citep{olympiadbench}.
\textit{Evaluation:} Pass@1 accuracy averaged over 8 independent samples.

\textbf{Medical reasoning.}
\textit{Models:} Qwen3-8B with math-trained E\textsuperscript{2}C-(SFT+RL) for zero-shot transfer, and Qwen3-8B / Llama3.1-8B for domain adaptation.
\textit{Benchmarks:} MedQA~\citep{medqa}, MedMCQA~\citep{medmcqa}, and six MMLU~\citep{MMLU} medical subsets.
\textit{Adaptation setup:} EF-SFT is trained on a 50k-sample exploration-only subset from ReasonMed with 10\% regularization data (10M tokens), compared against standard SFT on the full ReasonMed corpus (286M tokens).

\textbf{Test-time scaling.}
\textit{Model:} a single checkpoint, Qwen3-8B+E\textsuperscript{2}C-(SFT+RL).
\textit{Benchmark:} AIME'2024.
\textit{Protocol:} E\textsuperscript{2}C-Select, E\textsuperscript{2}C-ToT, and E\textsuperscript{2}C-ReAct Loop are compared with Self-Consistency, ToT, and ReAct at budgets $K$ or $N \in \{4, 8, 16, 32\}$. Sampled decoding uses temperature 0.9. We report Pass@1 accuracy and average tokens per question; full method details are in Appendix~\ref{app:tts_details}.
%\vspace{-2mm}
\subsection{Results}
%\vspace{-3mm}

\textbf{Mathematical Reasoning Benchmark Results} We conduct a sanity check comparing our E\textsuperscript{2}C models (Qwen3-8B/4B+E\textsuperscript{2}C-(SFT+RL)) against GRPO baselines, as shown in Table~\ref{tab:math_performance}. Our approach outperforms the baselines by 1.5\% (8B) and 1.9\% (4B), validating the effectiveness of the decomposition strategy. Notably, while paradigm shifts typically risk performance degradation, our method successfully maintains and enhances model capability through careful training design. The full E\textsuperscript{2}C framework ultimately surpasses the GRPO baseline by leveraging the decomposed structure, establishing a solid foundation for efficient test-time scaling.
 Training-stage ablations (Table~\ref{tab:math_performance}, bottom rows) are analyzed in Section~\ref{sec:ablations}.
%\vspace{-1mm}

\paragraph{Medical Reasoning Benchmark Results}
Table~\ref{tab:medical_generalization} presents the medical reasoning performance across three experimental settings. First, we establish competitive baselines by comparing against leading domain-specific 7B-8B models (HuatuoGPT-o1-7B~\citep{wang2024huatuogpt}, ReasonMed-7B~\citep{reasonmed}) and an open-source 14B medical LLM (Baichuan-M1-14B~\citep{baichuan}), with Qwen3-8B~\citep{qwen2025qwen3} serving as our base model reference.

For domain adaptation, we evaluate our EF-SFT approach (Section~\ref{sec:adaptation_and_inference}) against standard SFT on both Llama3.1-8B~\citep{dubey2024llama3} and Qwen3-8B architectures. As shown in Table~\ref{tab:medical_generalization}, EF-SFT achieves significant improvements of 3.9\% (Qwen3-8B) and 14.5\% (Llama3.1-8B) over standard SFT, while using only 3.5\% of the training tokens. The zero-shot transfer results further demonstrate that our mathematically-trained RL models attain performance comparable to specialized medical LLMs, validating the strong cross-domain generalization capability of our method.

% Medical
\begin{table*}[!t]
\centering
\caption{Performance Comparison of Models with Different Training Processes: Our inference paradigm demonstrates superior generalization, while EF-SFT shows improved efficiency and robustness. The six columns from Anatomy (AN), Clinical Knowledge (CK), College Biology (CB), College Medicine (CM), Medical Genetics (MG), and Professional Medicine (PM) are validation subsets of the MMLU benchmark. }
\label{tab:medical_generalization}
%\vspace{-2mm}
\resizebox{\textwidth}{!}{%
\begin{tabular}{@{}lcccccccccc@{}}
\toprule
\textbf{Model} & \textbf{MedQA} & \textbf{MedMCQA} & \textbf{AN} & \textbf{CK} & \textbf{CB} & \textbf{CM} & \textbf{MG} & \textbf{PM} & \textbf{\#Med-Tokens} & \textbf{Avg} \\ \midrule
\multicolumn{11}{l}{\textit{External Baselines}} \\
HuatuoGPT-o1-7B & \hlr{68.4} & \hlr{57.5} & \hlr{71.9} & \hlr{78.5} & \hly{88.2} & \hlr{67.6} & \hlr{80.0} & \hlr{77.6} & - & \hlr{73.7} \\
Baichuan-M1-14B & 76.5 & 65.2 & 77.3 & 83.6 & 87.9 & 80.7 & 89.1 & 88.8 & - & 81.1 \\
ReasonMed-7B & \hlr{66.9} & 65.1 & \hly{75.6} & \hlr{79.3} & \hlr{79.2} & \hlr{73.4} & \hly{85.0} & \hlr{80.9} & - & \hlr{75.7} \\
\midrule
\multicolumn{11}{l}{\textit{Our Method and Ablations}} \\
Qwen3-8B & \hly{71.4} & \hlr{59.5} & \hlr{68.0} & \hly{81.6} & \hlr{87.5} & \hly{78.0} & \hly{85.0} & \hly{83.8} & - & \hly{76.8} \\
Qwen3-8B + GRPO & \hly{74.0} & \hly{60.6} & \hly{75.0} & \hly{80.9} & 91.3 & \hly{81.8} & \hly{90.4} & \hlb{86.2} & - & \hly{79.1} \\
\textbf{Qwen3-8B+E\textsuperscript{2}C-(SFT+RL)} & \hlb{74.5} & \hlb{63.1} & \hlb{77.0} & \hlb{82.2} & \hlb{92.0} & \hlb{83.0} & \hlb{92.8} & 86.0 & - & \hlb{81.1} \\
\midrule
\multicolumn{11}{l}{\textit{SFT Models (using medical data)}} \\
Qwen3-8B + standard SFT & \hlr{58.2} & \hlr{52.3} & \hlr{68.8} & \hlr{80.8} & \hlr{89.0} & \hlr{73.7} & \hlr{83.3} & \hlr{79.0} & 286M & \hlr{73.1} \\
\textbf{Qwen3-8B + E\textsuperscript{2}C-SFT + EF-SFT} & \hlb{65.8} & \hlb{58.2} & \hlb{72.3} & \hlb{83.8} & \hlb{89.2} & \hlb{79.7} & \hlb{87.6} & \hlb{86.2} & 10M & \hlb{77.1} \\
Llama3.1-8B + ReasonMed SFT & \hlr{42.0} & \hlr{36.8} & \hlr{45.9} & \hlr{55.4} & \hlr{61.8} & \hlr{43.2} & \hlr{38.1} & \hlr{56.9} & 286M & \hlr{47.5} \\
\textbf{Llama3.1-8B + E\textsuperscript{2}C-SFT + EF-SFT} & \hlb{60.3} & \hlb{53.2} & \hlb{61.8} & \hlb{69.8} & \hlb{75.9} & \hlb{64.9} & \hlb{82.0} & \hlb{72.5} & 10M & \hlb{67.5} \\
\bottomrule
\end{tabular}%
}
\end{table*}

\begin{table*}[!t]
\centering
\caption{Test-Time Scaling Performance on AIME'2024 Benchmark (Qwen3-8B+E2C(SFT+RL)). Methods are grouped by scaling strategy: for each full-chain baseline, the corresponding E\textsuperscript{2}C variant (which concentrates additional compute on the short Exploration stage) is shown directly below. E\textsuperscript{2}C variants consistently achieve higher accuracy with substantially fewer tokens.}
\label{tab:scaling_comparison}
%\vspace{-2mm}
\resizebox{\textwidth}{!}{%
\begin{tabular}{@{}l|cc|cc|cc|cc@{}}
\toprule
& \multicolumn{2}{c|}{\textbf{Budget Level 1}} & \multicolumn{2}{c|}{\textbf{Budget Level 2}} & \multicolumn{2}{c|}{\textbf{Budget Level 3}} & \multicolumn{2}{c}{\textbf{Budget Level 4}} \\
\textbf{Method} & \textbf{Acc. (\%)} & \textbf{Tokens (k)} & \textbf{Acc. (\%)} & \textbf{Tokens (k)} & \textbf{Acc. (\%)} & \textbf{Tokens (k)} & \textbf{Acc. (\%)} & \textbf{Tokens (k)} \\ \midrule
Greedy CoT ($N=1$) & 36.7 & 2.2 & \multicolumn{6}{c}{\textit{(single decoding pass; N/A at higher budgets)}} \\
\midrule
\multicolumn{9}{l}{\textit{Multi-sample aggregation}} \\
Self-Consistency & 40.0 \tiny{(N=4)} & 11.1 & 40.0 \tiny{(N=8)} & 21.8 & 43.3 \tiny{(N=16)} & 43.7 & 50.0 \tiny{(N=32)} & 86.2 \\
E\textsuperscript{2}C-Select (Self LM-Judge) & 43.3 \tiny{(K=4)} & 3.4 & 26.7 \tiny{(K=8)} & 4.6 & 40.0 \tiny{(K=16)} & 6.5 & 40.0 \tiny{(K=32)} & 9.7 \\
\midrule
\multicolumn{9}{l}{\textit{Tree search}} \\
Tree-of-Thoughts (ToT) & 43.3 \tiny{(N=4)} & 13.6 & 43.3 \tiny{(N=8)} & 27.1 & 46.7 \tiny{(N=16)} & 43.4 & 50.0 \tiny{(N=32)} & 71.3 \\
E\textsuperscript{2}C-ToT & \hlb{46.7 \tiny{(K=4)}} & \hlb{9.9} & \hlb{46.7 \tiny{(K=8)}} & \hlb{18.3} & 43.3 \tiny{(K=16)} & \hlb{32.0} & \hlb{53.3 \tiny{(K=32)}} & \hlb{54.8} \\
\midrule
\multicolumn{9}{l}{\textit{Iterative refinement}} \\
ReAct ($N=1$) & 43.3 & 7.6 & \multicolumn{6}{c}{\textit{(single-loop reference; N/A at higher budgets)}} \\
E\textsuperscript{2}C-ReAct Loop & 43.3 \tiny{(K=4)} & \hlb{7.0} & 43.3 \tiny{(K=8)} & \hlb{5.7} & \hlb{46.7 \tiny{(K=16)}} & \hlb{7.8} & \hlb{53.3 \tiny{(K=32)}} & \hlb{12.4} \\
\bottomrule
\end{tabular}%
}
\end{table*}

\textbf{Test-Time Scaling Performance and Efficiency Analysis}$^{\ddagger}$\thispagestyle{ttspage}
Table~\ref{tab:scaling_comparison} shows a consistent pattern: for each full-chain baseline, the corresponding E\textsuperscript{2}C variant achieves higher accuracy at lower token cost by scaling only the Exploration stage. The strongest result is \textbf{E\textsuperscript{2}C-ReAct Loop} at $K{=}32$, which reaches \textbf{53.3\%} with only 12.4k tokens, outperforming Self-Consistency at $N{=}32$ (50.0\%, 86.2k tokens) and Tree-of-Thoughts (50.0\%, 71.3k tokens).

\begin{table}[t]
\centering
\caption{Ablation studies: (A) data construction strategy, (B) EF-SFT training steps, (C) regularization data mixing.}
\label{tab:ablation_and_entropy}
\resizebox{\columnwidth}{!}{
\begin{tabular}{@{} l c c @{}}
\toprule
\multicolumn{3}{@{}l}{\textit{Part A: SFT Data Construction Strategy}} \\
\midrule
\textbf{SFT Data Strategy} & \textbf{Plan-Guided} & \textbf{Plan-Free} \\
\midrule
Flawed (Reverse-Causal Summary) & 0.499 & 0.864 \\
\textbf{Proposed (Causal Generation)} & \hlb{0.998} & \hlb{1.000} \\
\bottomrule
\end{tabular}
}

\resizebox{\columnwidth}{!}{
\begin{tabular}{@{} l c c c c c @{}}
\toprule
\multicolumn{6}{@{}l}{\textit{Part B: EF-SFT Training Steps}} \\
\midrule
\textbf{Steps} & \textbf{GSM8K} & \textbf{Anatomy} & \textbf{CK} & \textbf{MedQA} & \textbf{MATH} \\
\midrule
SFT + 100 iter  & 83.0 & 70.4 & 80.6 & \hlb{66.4} & 70.5 \\
\textbf{SFT + 300 iter}  & 83.3 & \hlb{72.3} & \hlb{83.8} & 66.1 & \hlb{71.2} \\
SFT + 2000 iter & \hlb{84.0} & 69.8 & 78.4 & 63.5 & 71.1 \\
\bottomrule
\end{tabular}
}

\resizebox{\columnwidth}{!}{
\begin{tabular}{@{} l c c c c c @{}}
\toprule
\multicolumn{6}{@{}l}{\textit{Part C: Regularization Data Mixing}} \\
\midrule
\textbf{Setting} & \textbf{GSM8K} & \textbf{Anatomy} & \textbf{CK} & \textbf{MedQA} & \textbf{MATH} \\
\midrule
Medical Reg. ($\alpha{=}10\%$)  & 82.1 & \hlb{74.4} & 79.5 & 66.8 & 69.5 \\
\textbf{Math Reg. ($\alpha{=}10\%$)}    & \hlb{83.3} & 72.3 & \hlb{83.8} & 66.1 & \hlb{71.2} \\
Medical Reg. ($\alpha{=}100\%$) & 80.1 & 70.0 & 81.5 & \hlb{67.0} & 65.8 \\
No Reg. ($\alpha{=}0\%$)        & 82.1 & 68.2 & 77.8 & 63.5 & 70.8 \\
\bottomrule
\end{tabular}
}
\end{table}

\subsection{Ablations and Analysis}
\label{sec:ablations}
We ablate the main design choices behind E\textsuperscript{2}C across training, data construction, and domain adaptation (Table~\ref{tab:ablation_and_entropy}; bottom rows of Table~\ref{tab:math_performance}).

\paragraph{Training pipeline.}
On Qwen3-8B, prompt-only E\textsuperscript{2}C formatting reaches 46.6\% average accuracy, nearly matching zero-shot performance and showing that prompting alone cannot reliably instill the explore-execute interface. E\textsuperscript{2}C-SFT raises this to 46.8\%, confirming that explicit SFT is needed for the paradigm shift. Replacing generic GRPO with our E\textsuperscript{2}C-RL further improves 8B average accuracy from 57.7\% to 61.5\% (+3.8\%), with a similar +3.2\% gain on 4B, indicating that RL benefits come from exploiting decomposition rather than additional data alone.

\paragraph{Causal data construction (Part A).}
As discussed in Section~\ref{sec:dataset}, reverse-causal summarization lets the model learn to bypass exploration. Table~\ref{tab:ablation_and_entropy} quantifies this with \textit{Plan-Guided} adherence (execution follows its plan) and \textit{Plan-Free} adherence (execution answers without the plan). The flawed pipeline scores 0.499 and 0.864, respectively, whereas the causal pipeline in Algorithm~\ref{alg:correct_sft} reaches 0.998 and 1.000, enforcing the intended $p'(e\mid \pi,c)$ dependency required by Prop.~\ref{req:deterministic}.

\paragraph{EF-SFT training budget (Part B).}
Because EF-SFT updates only exploration segments, we ask how much domain-specific training is enough. On medical targets, 300 iterations ($\approx$5k samples) yields the best Anatomy (72.3\%) and Clinical Knowledge (83.8\%) scores. Extending to 2000 iterations improves in-domain GSM8K (84.0) but hurts MedQA (63.5 vs.\ 66.4 at 100 iterations), suggesting overfitting to exploration phrasing without improving transferable execution. We therefore use 300 iterations in all EF-SFT experiments.

\paragraph{Regularization mixing (Part C).}
Mixing 10\% base E\textsuperscript{2}C-SFT data during EF-SFT preserves math ability (GSM8K 83.3, MATH 71.2) while adapting medical reasoning. Removing regularization ($\alpha{=}0\%$) drops MedQA to 63.5, indicating that execution skills erode without replay data. Using the full exploration-execution sequence ($\alpha{=}100\%$) also underperforms, because it dilutes the exploration-only adaptation signal. Interestingly, math regularization at $\alpha{=}10\%$ outperforms medically matched regularization on average, so target-domain regularization data is not required for stable EF-SFT.

\section{Limitations and Future Works}
%\vspace{-3mm}

\paragraph{Error propagation through deterministic execution.}
Because the execution stage is designed to faithfully follow its conditioning plan, a flawed plan is not self-corrected during execution but instead carried out in full. This asymmetry means the framework's overall accuracy is disproportionately sensitive to plan-level errors, with no internal recovery mechanism within a single explore-execute pass.

\paragraph{Future Work.}
The explore-then-execute decomposition of E\textsuperscript{2}C maps naturally onto agentic settings. A direct extension is to treat the Exploration stage as a lightweight action-planning module that produces a structured sequence of tool calls or sub-tasks, while the Execution stage carries out each action step. This allows the plan-level scaling strategies developed here (selection, tree search, and iterative refinement) to be applied directly to multi-step tool-using agents.

\section{Related Work}

\textbf{From Chain-of-Thought to Structured Reasoning:} Chain-of-Thought (CoT) prompting~\citep{wei2022chain} significantly improves LLM reasoning, but its linear nature has motivated more robust structured paradigms that explore diverse reasoning paths~\citep{chen2025towards}. These include parallel sampling methods such as Self-Consistency~\citep{wang2022self,wan2025reasoning}, and more complex search structures including trees (ToT)~\citep{yao2023tree}, graphs (GoT)~\citep{besta2023graph,yao2024graph}, and forests (FoT)~\citep{bi2025forest}. While these paradigms expand the search space, they typically do not enforce a compact, trainable plan-to-execution interface within a single model. E\textsuperscript{2}C targets this gap through explicit phase separation and joint training.

\textbf{Planning and Decomposition in LLM Reasoning:} Planning before execution is already well studied in LLM reasoning, from subtask decomposition~\citep{zhou2022least, press2022measuring} and interleaved tool use~\citep{yao2022react, schick2023toolformer, patil2023gorilla} to LLM planners paired with external solvers~\citep{hao2023reasoning, liu2023llm} or multi-agent systems~\citep{yuan2024evoagent}. E\textsuperscript{2}C differs in scope and mechanism: it keeps both planning and execution inside one model, but separates them into structurally distinct phases with a compact plan interface, and trains that factorization directly via SFT and RL rather than treating it as a prompting convention. This design enables stable plan-level test-time scaling and exploration-only domain adaptation.

\textbf{Test-Time Scaling and Reasoning Efficiency:} Test-time scaling (TTS) improves performance by increasing inference-time compute~\citep{snell2024scaling, wu2025inference}. However, methods like Self-Consistency~\citep{wang2022self} and Tree-of-Thoughts~\citep{yao2023tree} generate multiple full-length reasoning chains, making their token cost grow proportionally with the budget. E\textsuperscript{2}C contributes a structurally motivated TTS strategy: by decoupling Exploration from Execution, it restricts search, selection, and refinement to the compact planning stage, so the token cost of scaling is determined by plan length rather than full solution length.

\section{Conclusion}

Through the proposed Explore-Execute Chain (E\textsuperscript{2}C), we present a reasoning framework that makes plan-then-execute structure explicit within a single LLM, enhancing efficiency and interpretability relative to undifferentiated CoT and prompt-only decomposition. The framework concentrates information in exploration, enabling domain adaptation with only 3.5\% of training tokens and a strong performance-cost trade-off on complex reasoning benchmarks. By restricting test-time scaling to compact plans, E\textsuperscript{2}C further offers a practical path toward efficient inference at higher compute budgets. We hope this decomposition encourages further work on trainable reasoning structure, where exploration and execution can be optimized, audited, and scaled independently within a unified model. Our results across mathematical and medical benchmarks, together with ablations on data construction and plan-level scaling, support explicit explore-execute factorization as a practical direction for both adaptation and inference efficiency.

\section*{Ethics Statement}
This work studies a reasoning framework, the \textbf{Explore–Execute Chain (E\textsuperscript{2}C)}, which separates lightweight exploratory sketches from a final execution step to improve efficiency, transparency, and controllability of LLM reasoning. Our experiments fine-tune and evaluate general-purpose LLMs on publicly available benchmarks (e.g., mathematics and domain reasoning datasets). We do not collect new human data, do not involve human or animal subjects, and do not process personally identifiable or sensitive information. Any third-party datasets used in this paper are publicly released for research purposes by their respective providers; we follow their licenses and usage terms. E\textsuperscript{2}C increases interpretability by exposing intermediate ``exploration'' traces, which can facilitate auditing and discourage over-reliance on hidden chain-of-thought. This study complies with the conference Code of Ethics.

\section*{Reproducibility Statement}
We have made extensive efforts to ensure the reproducibility of our work. All code used in this paper will be publicly released to facilitate independent verification and further research. We describe our experimental setup in Sec.~\ref{setup}. Detailed hyperparameters for training E\textsuperscript{2}C-SFT, E\textsuperscript{2}C-RL, GRPO, and EF-SFT are provided in Appendix~\ref{app:hyper}. Detailed setup for TTS experiments can be found in Appendix~\ref{app:tts_details}. We also include the prompt templates for data generation, the zero-shot prompt model, plan selection, execution, and exploration refinement in Appendix~\ref{prompt}.

\section*{Use of Large Language Models}
We utilized a large language model to enhance the language and clarity of our manuscript. Specifically, we employed Gemini 2.5 Flash with the following prompt to refine the initial draft: \textit{I am writing an academic paper in English. Please polish the following draft so that it adheres to the conventions of academic writing.}

\bibliography{references}

@inproceedings{wei2022chain,
  title={Chain-of-thought prompting elicits reasoning in large language models},
  author={Wei, Jason and Wang, Xuezhi and Schuurmans, Dale and Bosma, Maarten and Ichter, Brian and Xia, Fei and Chi, Ed and Le, Quoc and Zhou, Denny},
  booktitle={Advances in Neural Information Processing Systems},
  volume={35},
  pages={24824--24837},
  year={2022}
}

@inproceedings{wan2025reasoning,
    title = "Reasoning Aware Self-Consistency: Leveraging Reasoning Paths for Efficient {LLM} Sampling",
    author = "Wan, Guangya  and
      Wu, Yuqi  and
      Chen, Jie  and
      Li, Sheng",
    editor = "Chiruzzo, Luis  and
      Ritter, Alan  and
      Wang, Lu",
    booktitle = "Proceedings of the 2025 Conference of the Nations of the Americas Chapter of the Association for Computational Linguistics: Human Language Technologies (Volume 1: Long Papers)",
    month = apr,
    year = "2025",
    address = "Albuquerque, New Mexico",
    publisher = "Association for Computational Linguistics",
    url = "https://aclanthology.org/2025.naacl-long.184/",
    doi = "10.18653/v1/2025.naacl-long.184",
    pages = "3613--3635",
    ISBN = "979-8-89176-189-6"
}

@inproceedings{yao2024graph,
    title = "{G}o{T}: Effective Graph-of-Thought Reasoning in Language Models",
    author = "Yao, Yao  and
      Li, Zuchao  and
      Zhao, Hai",
    editor = "Duh, Kevin  and
      Gomez, Helena  and
      Bethard, Steven",
    booktitle = "Findings of the Association for Computational Linguistics: NAACL 2024",
    month = jun,
    year = "2024",
    address = "Mexico City, Mexico",
    publisher = "Association for Computational Linguistics",
    url = "https://aclanthology.org/2024.findings-naacl.183/",
    doi = "10.18653/v1/2024.findings-naacl.183",
    pages = "2901--2921"
}

@inproceedings{
wu2025inference,
title={Inference Scaling Laws: An Empirical Analysis of Compute-Optimal Inference for {LLM} Problem-Solving},
author={Yangzhen Wu and Zhiqing Sun and Shanda Li and Sean Welleck and Yiming Yang},
booktitle={The Thirteenth International Conference on Learning Representations},
year={2025},
url={https://openreview.net/forum?id=VNckp7JEHn}
}

@inproceedings{zhou2022least,
  title={Least-to-most prompting enables complex reasoning in large language models},
  author={Zhou, Denny and Sch{\"a}rli, Nathanael and Hou, Le and Wei, Jason and Scales, Nathan and Wang, Xuezhi and Schuurmans, Dale and Cui, Claire and Bousquet, Olivier and Le, Quoc and Chi, Ed},
  booktitle={International Conference on Learning Representations},
  year={2023}
}

@article{press2022measuring,
  title={Measuring and narrowing the compositionality gap in language models},
  author={Press, Ofir and Yoran, Or and Schick, Timo and Schmid, Idan and Fisch, Ayal and Goldberg, Yoav and Misra, Kanishka},
  journal={arXiv preprint arXiv:2210.03350},
  year={2022}
}

@article{schick2023toolformer,
  title={Toolformer: Language models can teach themselves to use tools},
  author={Schick, Timo and Dwivedi-Yu, Jane and Dessi, Roberto and Raileanu, Roberta and Tsvigun, Maria and Cances, Gautier and Smaili, Najma},
  journal={arXiv preprint arXiv:2302.04761},
  year={2023}
}

@article{patil2023gorilla,
  title={Gorilla: Large language model connected with massive apis},
  author={Patil, Shishir G and Zhang, Tianjun and Wang, Xin and Gonzalez, Joseph E},
  journal={arXiv preprint arXiv:2305.15334},
  year={2023}
}

@article{hao2023reasoning,
  title={Reasoning with language model is planning with world model},
  author={Hao, Shibo and Gu, Yi and Ma, Haodi and Hong, Joshua and Wang, Zhen and Wang, Daisy and Hu, Zhiting},
  journal={arXiv preprint arXiv:2305.14992},
  year={2023}
}

@article{liu2023llm,
  title={LLM+ P: Empowering large language models with optimal planning proficiency},
  author={Liu, Boi-Faltings and Liu, Zhang-Wei and Jiang, Ruibo and Lyu, Yisong and Du, Yizhou and Wu, F. and Liu, Yu-Feng},
  journal={arXiv preprint arXiv:2304.11477},
  year={2023}
}

@article{rafailov2023direct,
  title={Direct preference optimization: Your language model is secretly a reward model},
  author={Rafailov, Rafael and Sharma, Archit and Mitchell, Eric and Ermon, Stefano and Manning, Christopher D and Finn, Chelsea},
  journal={arXiv preprint arXiv:2305.18290},
  year={2023}
}

@misc{wang2022self,
      title={Self-Consistency Improves Chain of Thought Reasoning in Language Models}, 
      author={Xuezhi Wang and Jason Wei and Dale Schuurmans and Quoc Le and Ed Chi and Sharan Narang and Aakanksha Chowdhery and Denny Zhou},
      year={2022},
      eprint={2203.11171},
      archivePrefix={arXiv},
      primaryClass={cs.CL}
}

@inproceedings{yao2023tree,
  title={Tree of Thoughts: Deliberate Problem Solving with Large Language Models},
  author={Yao, Shunyu and Yu, Dian and Zhao, Jeffrey and Sha, Izhak and Savarese, Silvio and an, Tao},
  booktitle={Advances in Neural Information Processing Systems},
  year={2023}
}

@misc{besta2023graph,
      title={Graph of Thoughts: Solving Elaborate Problems with Large Language Models}, 
      author={Maciej Besta and Nils Blach and Ales Kubicek and Robert Gerstenberger and Lukas Gianinazzi and Kedar Tatwawadi and Joana Einsiedler and Daria Costanzo and Gregor J. Räbsamen and Michael Wand and Hermann sundry and Torsten Hoefler},
      year={2023},
      eprint={2308.09687},
      archivePrefix={arXiv},
      primaryClass={cs.CL}
}

@misc{shinn2023reflexion,
      title={Reflexion: Language Agents with Verbal Reinforcement Learning}, 
      author={Noah Shinn and Federico Cassano and Edward Berman and Ashwin Gopinath and Karthik Narasimhan and Shunyu Yao},
      year={2023},
      eprint={2303.11366},
      archivePrefix={arXiv},
      primaryClass={cs.CL}
}

@misc{yao2022react,
      title={ReAct: Synergizing Reasoning and Acting in Language Models}, 
      author={Shunyu Yao and Jeffrey Zhao and Dian Yu and Nan Du and Izhak Shafran and Karthik Narasimhan and Yuan Cao},
      year={2022},
      eprint={2210.03629},
      archivePrefix={arXiv},
      primaryClass={cs.CL}
}

@misc{bi2025forest,
      title={Forest-of-Thought: Scaling Test-Time Compute for Enhancing LLM Reasoning}, 
      author={Zhenni Bi and Kai Han and Chuanjian Liu and Yehui Tang and Yunhe Wang},
      year={2025},
      eprint={2412.09078},
      archivePrefix={arXiv},
      primaryClass={cs.CL}
}

@misc{liao2025fractured,
      title={Fractured Chain-of-Thought Reasoning}, 
      author={Baohao Liao and Hanze Dong and Yuhui Xu and Doyen Sahoo and Christof Monz and Junnan Li and Caiming Xiong},
      year={2025},
      eprint={2505.12992},
      archivePrefix={arXiv},
      primaryClass={cs.LG}
}

@misc{xu2025softcotpp,
      title={SoftCoT++: Test-Time Scaling with Soft Chain-of-Thought Reasoning}, 
      author={Yige Xu and Xu Guo and Zhiwei Zeng and Chunyan Miao},
      year={2025},
      eprint={2505.11484},
      archivePrefix={arXiv},
      primaryClass={cs.CL}
}

@misc{xu2025adaptive,
      title={Adaptive Termination for Multi-round Parallel Reasoning: An Universal Semantic Entropy-Guided Framework}, 
      author={Zenan Xu and Zexuan Qiu and Guanhua Huang and Kun Li and Siheng Li and Chenchen Zhang and Kejiao Li and Qi Yi and Yuhao Jiang and Bo Zhou and Fengzong Lian and Zhanhui Kang},
      year={2025},
      eprint={2507.06012},
      archivePrefix={arXiv},
      primaryClass={cs.CL}
}

@misc{li2025compressing,
      title={Compressing Chain-of-Thought in LLMs via Step Entropy}, 
      author={Zeju Li and Jianyuan Zhong and Ziyang Zheng and Xiangyu Wen and Zhijian Xu and Yingying Cheng and Fan Zhang and Qiang Xu},
      year={2025},
      eprint={2508.03346},
      archivePrefix={arXiv},
      primaryClass={cs.AI}
}

@misc{zheng2025fr3e,
      title={First Return, Entropy-Eliciting Explore}, 
      author={Tianyu Zheng and Tianshun Xing and Qingshui Gu and Taoran Liang and Xingwei Qu and Xin Zhou and Yizhi Li and Zhoufutu Wen and Chenghua Lin and Wenhao Huang and Qian Liu and Ge Zhang and Zejun Ma},
      year={2025},
      eprint={2507.07017},
      archivePrefix={arXiv},
      primaryClass={cs.AI}
}

@misc{gan2025cotspace,
      title={{CoT-Space}: A Theoretical Framework for Internal Slow-Thinking via Reinforcement Learning}, 
      author={Zeyu Gan and Hao Yi and Yong Liu},
      year={2025},
      eprint={2509.04027},
      archivePrefix={arXiv},
      primaryClass={cs.AI}
}

@misc{snell2024scaling,
    title={Scaling LLM Test-Time Compute Optimally Can Be More Effective Than Scaling Model Parameters}, 
    author={Charlie Snell and Jaehoon Lee and Kelvin Xu and Aviral Kumar},
    year={2024},
    eprint={2408.03314},
    archivePrefix={arXiv},
    primaryClass={cs.LG}
}

@misc{chen2025towards,
    title={Towards Reasoning Era: A Survey of Long Chain-of-Thought for Reasoning in Large Language Models}, 
    author={Qiguang Chen and Libo Qin and Jinhao Liu and Dengyun Peng and Jiannan Guan and Peng Wang and Mengkang Hu and Yuhang Zhou and Te Gao and Wanxiang Che},
    year={2025},
    eprint={2503.09567},
    archivePrefix={arXiv},
    primaryClass={cs.CL}
}

@misc{yuan2024evoagent,
    title={EvoAgent: Towards Automatic Multi-Agent Generation via Evolutionary Algorithms}, 
    author={Siyuan Yuan and Kairui Song and Jia-Hao Chen and Xiao-Hui Tan and Dian-Hui Li and Dong-Sheng Yang},
    year={2024},
    eprint={2405.16510},
    archivePrefix={arXiv},
    primaryClass={cs.AI}
}

@article{achtziger2007rubicon,
  title={Rubicon model of action phases},
  author={Achtziger, Anja and Gollwitzer, Peter M},
  year={2007}
}

@misc{wang2024huatuogpt,
    title={{HuatuoGPT}, a General-purpose Chinese Medical Large Language Model}, 
    author={Junying Wang and Zhaonan Li and Renfeng Pu and Saijiang Shi and Yitong Meng and Zhaokun Wang and Yixin Liu and Jianing Zhou and Wenjia Zhang and Jialiang Chen and Yefeng Zheng and Hong-Yin Mey},
    year={2024},
    eprint={2405.18524},
    archivePrefix={arXiv},
    primaryClass={cs.CL}
}

@article{reasonmed,
  title={ReasonMed: A 370K Multi-Agent Generated Dataset for Advancing Medical Reasoning},
  author={Sun, Yu and Qian, Xingyu and Xu, Weiwen and Zhang, Hao and Xiao, Chenghao and Li, Long and Rong, Yu and Huang, Wenbing and Bai, Qifeng and Xu, Tingyang},
  journal={arXiv preprint arXiv:2506.09513},
  year={2025}
}

@article{yu2025dapo,
  title={Dapo: An open-source llm reinforcement learning system at scale},
  author={Yu, Qiying and Zhang, Zheng and Zhu, Ruofei and Yuan, Yufeng and Zuo, Xiaochen and Yue, Yu and Dai, Weinan and Fan, Tiantian and Liu, Gaohong and Liu, Lingjun and others},
  journal={arXiv preprint arXiv:2503.14476},
  year={2025}
}

@misc{qwen2025qwen3,
    title={{Qwen3} Technical Report}, 
    author={An Yang and Anfeng Li and Baosong Yang and Beichen Zhang and Binyuan Hui and Bo Zheng and Bowen Yu and Chang Gao and Chengen Huang and Chenxu Lv and et al.},
    year={2025},
    eprint={2505.09388},
    archivePrefix={arXiv},
    primaryClass={cs.CL}
}

@article{adam,
  title={Adam: A method for stochastic optimization},
  author={Kingma, Diederik P},
  journal={arXiv preprint arXiv:1412.6980},
  year={2014}
}

@misc{dubey2024llama3,
    title={The Llama 3 Herd of Models}, 
    author={Abhimanyu Dubey and Abhinav Jauhri and Abhinav Pandey and Abhishek Kadian and Ahmad Al-Dahle and Aiesha Letman and Akhil Mathur and Alan Schelten and Amy Yang and Angela Fan and et al.},
    year={2024},
    eprint={2407.21783},
    archivePrefix={arXiv},
    primaryClass={cs.CL}
}

@article{wang2025emergent,
  title={Emergent hierarchical reasoning in llms through reinforcement learning},
  author={Wang, Haozhe and Xu, Qixin and Liu, Che and Wu, Junhong and Lin, Fangzhen and Chen, Wenhu},
  journal={arXiv preprint arXiv:2509.03646},
  year={2025}
}

@inproceedings{medmcqa,
  title={Medmcqa: A large-scale multi-subject multi-choice dataset for medical domain question answering},
  author={Pal, Ankit and Umapathi, Logesh Kumar and Sankarasubbu, Malaikannan},
  booktitle={Conference on health, inference, and learning},
  pages={248--260},
  year={2022},
  organization={PMLR}
}

@article{medqa,
  title={What disease does this patient have? a large-scale open domain question answering dataset from medical exams},
  author={Jin, Di and Pan, Eileen and Oufattole, Nassim and Weng, Wei-Hung and Fang, Hanyi and Szolovits, Peter},
  journal={Applied Sciences},
  volume={11},
  number={14},
  pages={6421},
  year={2021},
  publisher={MDPI}
}

@article{MMLU,
  title={Measuring massive multitask language understanding},
  author={Hendrycks, Dan and Burns, Collin and Basart, Steven and Zou, Andy and Mazeika, Mantas and Song, Dawn and Steinhardt, Jacob},
  journal={arXiv preprint arXiv:2009.03300},
  year={2020}
}

@article{baichuan,
  title={Baichuan-M1: Pushing the Medical Capability of Large Language Models},
  author={Wang, Bingning and Zhao, Haizhou and Zhou, Huozhi and Song, Liang and Xu, Mingyu and Cheng, Wei and Zeng, Xiangrong and Zhang, Yupeng and Huo, Yuqi and Wang, Zecheng and Zhao, Zhengyun and others},
  journal={arXiv preprint arXiv:2502.12671},
  year={2025}
}

@article{math,
  title={Measuring mathematical problem solving with the math dataset, 2021},
  author={Hendrycks, Dan and Burns, Collin and Kadavath, Saurav and Arora, Akul and Basart, Steven and Tang, Eric and Song, Dawn and Steinhardt, Jacob},
  journal={URL https://arxiv. org/abs/2103.03874},
  volume={2},
  year={2024}
}

@article{olympiadbench,
  title={Olympiadbench: A challenging benchmark for promoting agi with olympiad-level bilingual multimodal scientific problems},
  author={He, Chaoqun and Luo, Renjie and Bai, Yuzhuo and Hu, Shengding and Thai, Zhen Leng and Shen, Junhao and Hu, Jinyi and Han, Xu and Huang, Yujie and Zhang, Yuxiang and others},
  journal={arXiv preprint arXiv:2402.14008},
  year={2024}
}

@misc{math500,
  title        = {MATH-500},
  author       = {{HuggingFaceH4}},
  howpublished = {\url{https://huggingface.co/datasets/HuggingFaceH4/MATH-500}},
  year         = {2025},
  note         = {Hugging Face Dataset}
}

@article{verl,
  title   = {HybridFlow: A Flexible and Efficient RLHF Framework},
  author  = {Guangming Sheng and Chi Zhang and Zilingfeng Ye and Xibin Wu and Wang Zhang and Ru Zhang and Yanghua Peng and Haibin Lin and Chuan Wu},
  year    = {2024},
  journal = {arXiv preprint arXiv: 2409.19256}
}

@misc{openr1math,
  title        = {Open R1: A fully open reproduction of DeepSeek-R1},
  author       = {{deepseek}},
  howpublished = {\url{https://github.com/huggingface/open-r1}},
  month        = {January},
  year         = {2025}
}

\appendix
\section{Appendix}
% Cognitive Model Analysis
\subsection{Cognitive Model Analysis}
In this section, we analyze cognitive models to derive high-level design insights for our method.

\subsubsection{Rubicon Model of Action Phases}
The \textbf{Rubicon Model of Action Phases}~\citep{achtziger2007rubicon}, proposed by \textbf{Heckhausen} and \textbf{Gollwitzer}, provides a framework for how individuals prepare for and pursue goals. It divides goal pursuit into four stages: \textbf{goal setting}, \textbf{planning}, \textbf{action}, and \textbf{evaluation}.

\begin{enumerate}
\item \textbf{Goal Setting}: Individuals identify and adopt a goal, motivated by a need or desire.
\item \textbf{Planning}: After a goal is adopted, individuals generate strategies to achieve it and assess their potential effectiveness.
\item \textbf{Action}: Once a strategy is selected, the individual commits to it and executes it. Crossing the “Rubicon” marks this commitment and the transition to action.
\item \textbf{Evaluation}: Outcomes are assessed to inform adjustments to subsequent plans or actions.
\end{enumerate}

A key contribution of the Rubicon Model is the sharp distinction between planning and execution. After commitment (crossing the Rubicon), attention is devoted to execution rather than continued exploration or second-guessing. This separation mitigates cognitive overload that could arise from ongoing re-evaluation during task execution.

\subsubsection{Connecting E\textsuperscript{2}C with the Method}
We formally express E\textsuperscript{2}C as
\begin{equation}
\resizebox{\columnwidth}{!}{$\displaystyle
\underbrace{p(e \mid c)}_{\text{Coupled Reasoning Process}} \rightarrow
\underbrace{p'(\pi, e \mid c)}_{\text{Explore-Execute Chain}} =
\underbrace{p'(\pi \mid c)}_{\textcolor{red}{\text{Highly Informative}}} \cdot
\underbrace{p'(e \mid \pi, c)}_{\textcolor{blue}{\text{Highly Deterministic}}}$}
\end{equation}

\noindent
\textbf{\(p'(\pi \mid c)\) as the Planning Phase}: In the Rubicon framework, planning entails generating candidate strategies. Analogously, in E\textsuperscript{2}C, \(p'(\pi \mid c)\) produces multiple candidate plans \(\pi\) from context \(c\). These plans are highly informative, capturing the critical information needed to solve the task. This exploration corresponds to the goal-setting and planning stages, where alternatives are considered before selection.

\noindent
\textbf{\(p'(e \mid \pi, c)\) as the Execution Phase}: Once plans are available, E\textsuperscript{2}C transitions to execution. The distribution \(p'(e \mid \pi, c)\) reflects a highly deterministic process that follows the selected plan \(\pi\) under context \(c\). This mirrors the action phase of the Rubicon Model: the agent executes the committed plan without revisiting discarded alternatives.

Thus, the separation between \(p'(\pi \mid c)\) and \(p'(e \mid \pi, c)\) in E\textsuperscript{2}C parallels the explore–then–execute dynamics of the Rubicon Model: first enumerate options, then execute deterministically.

\subsubsection{Cognitive and Computational Efficiency}
Separating exploration from execution confers efficiency benefits in both cognition and computation. Cognitively, once commitment occurs, resources are focused on carrying out the chosen plan without distraction from alternatives. Computationally, E\textsuperscript{2}C avoids the overhead of re-evaluating multiple plans during execution. The deterministic execution phase concentrates compute on following the selected plan, yielding faster and more reliable performance than continually interleaving exploration with action.

\subsubsection{Interpretability and Transparency}
The exploration–execution split also improves \textbf{interpretability}. In the Rubicon Model, one can explain an action by the plan selected during the planning stage. Likewise, E\textsuperscript{2}C makes the reasoning path explicit: multiple candidate plans are generated (exploration), and one is chosen and followed (execution). This transparency further supports \textbf{scalability}: the exploration component can be adapted to new tasks and domains, while the execution component remains stable, enabling flexible and extensible reasoning across settings.

% training details important reproduct

\subsection{The Details of The Experiments}
In this section, we introduce the details of our main experiments in the main paper for reproducibility purposes, including the detailed hyperparameter settings and the reward designs.
\subsubsection{Hyperparameter Settings}
\label{app:hyper}
\paragraph{E\textsuperscript{2}C-SFT and EF-SFT Training}

For both E\textsuperscript{2}C-SFT and EF-SFT training, the hyperparameters are summarized in Table~\ref{hyp:sft}:

\begin{table*}[t]
\centering
\resizebox{0.55\textwidth}{!}{%
\begin{tabular}{|c|c|}
\hline
\textbf{Hyperparameter} & \textbf{Value} \\
\hline
Learning Rate & \(1.0 \times 10^{-5}\) \\
Optimizer & Adam\citep{adam} (\( \beta_1 = 0.9, \beta_2 = 0.95 \)) \\
Weight Decay & 0.01 \\
Learning Rate Scheduler & Cosine with 10\% warmup ratio \\
Batch Size & 160 \\
Micro-batch Size per GPU & 20 \\
Gradient Clipping & 1.0 \\
Total Epochs & 1 \\
\hline
\end{tabular}
}
\caption{Hyperparameters for E\textsuperscript{2}C-SFT and EF-SFT Training}
\label{hyp:sft}
\end{table*}

\paragraph{E\textsuperscript{2}C-RL and GRPO Training}
The hyperparameters for E\textsuperscript{2}C-RL and GRPO training are summarized in Table~\ref{hyp:RL}, where the experiments include E\textsuperscript{2}C Stage 1 (E2C-stg1), E\textsuperscript{2}C Stage 2 (E2C-stg2), and GRPO:

\begin{table*}[t]
\centering
\resizebox{0.95\textwidth}{!}{%
\begin{tabular}{|c|c|c|c|}
\hline
\textbf{Hyperparameter} & \textbf{E\textsuperscript{2}C-stage1} & \textbf{E\textsuperscript{2}C-stage2} & \textbf{GRPO} \\
\hline
Batch Size & 256 & 256 & 128 \\
Overlong Buffer Length & 4096 & 4096 & 4096 \\
Maximum Response Length & 8192 & 8192  & 8192 \\
Learning Rate & \(1.0 \times 10^{-6}\) & \(1.0 \times 10^{-6}\) & \(1.0 \times 10^{-6}\) \\
Mini-batch Size for GRPO Updates & 32 & 32 & 32 \\
KL Loss Coefficient \( \beta \) & 0.001  & 0 & 0 \\
Rollout Number $k$& 32 & 8 & 8 \\
Temperature & 1.3 & 1.0 & 1.0 \\
Training Epochs & 1 & 1 & 5 \\
Clip ratio ($\varepsilon$)& 0.2 & 0.2 & 0.2 \\
\hline
\end{tabular}
}
\caption{Hyperparameters for E\textsuperscript{2}C-RL and GRPO Training}
\label{hyp:RL}
\end{table*}

\subsubsection{Reward Details for RL training}
\paragraph{Format Reward Calculation for E\textsuperscript{2}C Training}

For the \textbf{E\textsuperscript{2}C} training, the format reward consists of two components: the length reward and the instruction reward. These rewards are computed as follows:

\textbf{Length Reward}: This reward measures how well the output length matches the expected length. It is computed as:

\[
r_l = -\text{clip}\left(0,1,\frac{L - L_{valid}}{L_{buffer}}\right)
\]

where:
\(L\) is the length of the generated output;
\(L_{valid}\) is the length of the valid portion of the response;
\(L_{buffer}\) is the overlong buffer length.

\textbf{Instruction Reward}: The instruction reward is specific to the \textbf{E\textsuperscript{2}C} model and is added to the reward function when it comes to E\textsuperscript{2}C model. This reward measures the alignment between the instructions generated during the exploration phase and the execution phase. It is computed by extracting the step titles from both the exploration and execution phases using regular expressions. Denote these sets of instructions as \( S_1 \) (exploration) and \( S_2 \) (execution). The instruction reward is defined as:

\[
r_{\text{instr}} = 0.1*(\frac{|S_1 \cap S_2|}{\max(|S_1|, |S_2|)}-1)
\]

where:
\( S_1 \) is the set of instructions generated during the exploration phase;
\( S_2 \) is the set of instructions generated during the execution phase;
\( |S_1 \cap S_2| \) is the intersection of the sets \( S_1 \) and \( S_2 \);
\( \max(|S_1|, |S_2|) \) is the maximum size of the two sets.

The instruction reward incentivizes the model to generate instructions that align well between the exploration and execution phases, encouraging consistency. This reward is crucial for \textbf{E\textsuperscript{2}C} models to ensure that the reasoning process is coherent between the exploration and execution stages.

\paragraph{Format Reward Calculation for GRPO Training}

For \textbf{GRPO} training, the format reward is simpler and consists solely of the length reward, which is computed using the same formula as in E\textsuperscript{2}C:

\[
r_l = -\text{clip}\left(0,1,\frac{\text{length}_{\text{output}} - \text{valid}_{\text{length}}}{\text{buffer}_{\text{length}}}\right)
\]

In GRPO, no instruction reward is applied, and the focus is entirely on the length of the response, ensuring that the output adheres to the expected length constraints.

% two algorithm, maybe need to be merged

\subsection{Details of the Algorithm}
\paragraph{Algorithm of E\textsuperscript{2}C-SFT Data Generation}
Algorithm~\ref{alg:correct_sft} formalizes the causal data pipeline introduced in Section~\ref{sec:dataset}.
\begin{algorithm}
		\caption{E\textsuperscript{2}C-SFT Data Generation}
        \label{alg:correct_sft}
		\begin{algorithmic}[1]
			\State $\mathcal{D}_{\text{synth}} \gets \emptyset$
			\For{each question $q$}
            \State $\text{solution} \gets \text{Model}_{\text{base}}(q)$
			\State $\text{exploration} \gets \text{Summarize}(\text{solution})$
			\State $\text{prompt} \gets$
            \parbox[t]{0.78\linewidth}{``Given question: $q$.
            Follow exploration: exploration. Execute step-by-step:''}
			\State $\text{execution} \gets \text{Model}_{\text{base}}(\text{prompt})$
			\State $\mathcal{D}_{\text{synth}} \gets \mathcal{D}_{\text{synth}} \cup \{(q, \text{(exploration,execution)})\}$
			\EndFor
			\State \Return $\mathcal{D}_{\text{synth}}$
		\end{algorithmic}
	\end{algorithm}
\paragraph{Algorithm of Exploration-Focused SFT (EF-SFT) Data Generation}
Algorithm~\ref{alg:EF-SFT} is a formal and detailed description for EF-SFT Data Generation.
\begin{algorithm}
    \caption{EF-SFT Data Generation}
    \label{alg:EF-SFT}
    \begin{algorithmic}[1]
        \Require Base E\textsuperscript{2}C dataset $\mathcal{D}_{\text{base}}$
        \Require Domain-specific dataset $\mathcal{D}_{\text{domain}}$
        \Require Mixing ratio $\alpha \in [0, 1]$,Target Dataset size $n_{\text{target}}$
        \Ensure EF-SFT training dataset $\mathcal{D}_{\text{EF-SFT}}$
        
        \State $\mathcal{D}_{\text{explore}} \gets \emptyset$
        \For{each example $(q, a) \in \mathcal{D}_{\text{domain}}$}
            \State Extract exploration segment: $e \gets \text{ExtractExploration}(a)$
            \State $\mathcal{D}_{\text{explore}} \gets \mathcal{D}_{\text{explore}} \cup \{(q, e)\}$
        \EndFor
        \State $n_{\text{base}} \gets \alpha \times n_{\text{target}}$ \Comment{$\alpha\%$ from base dataset}
        \State $n_{\text{explore}} \gets (1 - \alpha) \times n_{\text{target}}$ \Comment{$(1-\alpha)\%$ from exploration data}
        
        \State $\mathcal{D}_{\text{base}}^{\text{sub}} \gets \text{Subsample}(\mathcal{D}_{\text{base}}, n_{\text{base}})$
        \State $\mathcal{D}_{\text{explore}}^{\text{sub}} \gets \text{Subsample}(\mathcal{D}_{\text{explore}}, n_{\text{explore}})$
        
        \State $\mathcal{D}_{\text{EF-SFT}} \gets \mathcal{D}_{\text{base}}^{\text{sub}} \cup \mathcal{D}_{\text{explore}}^{\text{sub}}$
        \State \Return $\mathcal{D}_{\text{EF-SFT}}$
    \end{algorithmic}
\end{algorithm}
\paragraph{Algorithm of E\textsuperscript{2}C Test-Time Scaling}
Algorithm~\ref{alg:tts} formalizes the shared explore-select-execute-refine template introduced in Section~\ref{sec:adaptation_and_inference}.
\begin{algorithm}
\caption{E\textsuperscript{2}C Test-Time Scaling}
\label{alg:tts}
\begin{algorithmic}[1]
\State Generate or expand $K$ exploration segments: $\{\pi_1, \pi_2, \dots, \pi_K\}$
\State Select a compact candidate set $\Pi^*$ via judge, clustering, tree pruning, or fallback plans
\For{each selected exploration $\pi_i^* \in \Pi^*$}
\State Generate deterministic execution: $a_i \gets \text{Execute}(\pi_i^*)$
\If{execution stalls and the budget remains}
\State Revise or resample the exploration: $\pi_i^* \gets \text{Refine}(\pi_i^*)$
\State Retry execution with the revised exploration
\EndIf
\EndFor
\State Let $a_{\text{final}}$ be the selected answer or the weighted-vote winner
\State \Return $a_{\text{final}}$
\end{algorithmic}
\end{algorithm}

% Test time details
\subsection{Test-Time Scaling Experimental Details}
\label{app:tts_details}
This section provides a detailed description of the experimental setup for the test-time scaling comparison presented in Table~\ref{tab:scaling_comparison}, following the template in Algorithm~\ref{alg:tts}.
\paragraph{Objective and General Setup}
The primary goal was to test whether E\textsuperscript{2}C methods that scale the Exploration stage outperform full-chain baselines that scale complete reasoning chains. All methods used the same checkpoint (\textbf{Qwen3-8B+E\textsuperscript{2}C-(SFT+RL)}) to ensure a fair comparison of inference strategies. Diverse sampling steps used temperature 0.9. Performance is reported as Pass@1 accuracy, and cost is measured by the average total generated tokens per question.

\paragraph{Full-Chain Baselines}
\begin{itemize}[leftmargin=15pt,topsep=3pt, itemsep=1pt]
    \item \textbf{Greedy CoT ($N{=}1$)}: A single reasoning chain generated with greedy decoding. Serves as the zero-budget reference.
    \item \textbf{Self-Consistency (SC)}: For each budget level $N \in \{4, 8, 16, 32\}$, $N$ full independent CoT chains were sampled and majority-voted.
    \item \textbf{Tree-of-Thoughts (ToT)}: ToT tree expansion was applied to full reasoning chains. Branch factor 2, depth $\lceil\log_2 N\rceil$; leaf nodes were completed to full solutions and majority-voted.
    \item \textbf{ReAct ($N{=}1$)}: A single-pass ReAct-style attempt that interleaves reasoning with self-correction; serves as a single-loop iterative reference.
\end{itemize}

\paragraph{E\textsuperscript{2}C Methods}
All E\textsuperscript{2}C variants use the model's Exploration/Execution format; $K \in \{4, 8, 16, 32\}$ controls the exploration budget.
\begin{itemize}[leftmargin=15pt,topsep=3pt, itemsep=1pt]
    \item \textbf{E\textsuperscript{2}C-Select (Self LM-Judge)}: $K$ exploration plans were sampled; the model then acted as a judge to select the single most promising plan; one deterministic execution was generated from that plan.
    \item \textbf{E\textsuperscript{2}C-ToT}: ToT-style branching was applied solely over the Exploration stage (branch factor 2, depth $\lceil\log_2 K\rceil$). The resulting plan leaves were clustered with K-Means (M=3 clusters; \texttt{all-mpnet-base-v2} embeddings); the centroid plan of each cluster was executed and answers were aggregated by cluster-size-weighted majority vote.
    \item \textbf{E\textsuperscript{2}C-ReAct Loop}: $K$ explorations were sampled; the self LM-judge selected an initial plan; it was executed deterministically. If no valid boxed answer was found or the execution showed looping/hallucination patterns, the model revised the plan using the failed execution as feedback (or switched to an unused candidate plan). The loop was capped at 3 refinement rounds; each refinement call used at most 1024 tokens.
\end{itemize}

% visualization
\subsection{Entropy Visualization of Different RL Settings and Analysis}
\label{entropy}
\begin{figure}[t]
    \centering
    \includegraphics[width=\columnwidth]{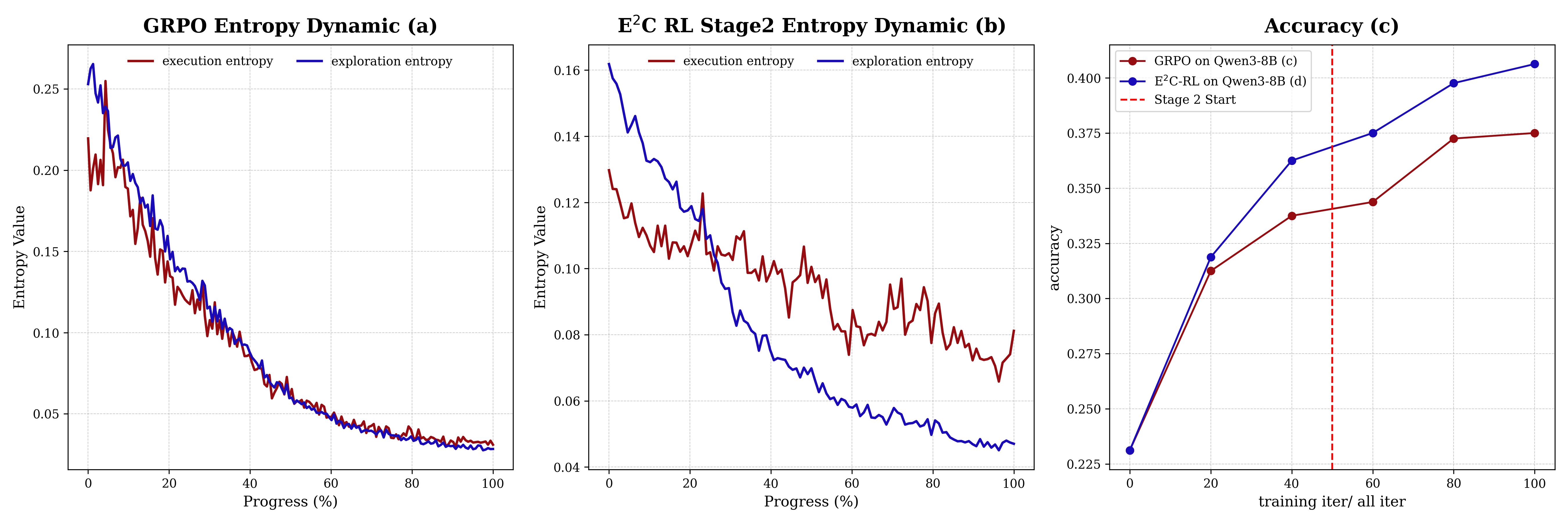}
    \caption{
        A comparison of training dynamics on the AIME'24 benchmark.
        The application of our token-weighting coefficient $\lambda_{i,t}$ (b) facilitates faster entropy reduction and superior performance improvement compared to the baseline without it (a).
    }
    \label{fig:combine}
\end{figure}

In this part, we visualize the entropy dynamics and the accuracy on the AIME'24 benchmark during RL training. The results demonstrate that applying our token-weighting coefficient $\lambda_{i,t}$ to exploration tokens facilitates a rapid drop in entropy and a better performance improvement, as shown in \textbf{Fig.~\ref{fig:combine}}. This is achieved by effectively amplifying high-quality plans while suppressing poor ones.

% Prompt
\subsection{Prompt Details}
\label{prompt}
\paragraph{E\textsuperscript{2}C-SFT Dataset Construction prompt}

\subparagraph*{Exploration Phase Prompt}
The following prompt is used to extract the high-level exploration plan from the reasoning process:

\begin{quote}
\noindent\textbf{Role:} You are an expert problem-solver.\\
\textbf{Task:} Distill a complex reasoning process into a clear, actionable plan.\\

\noindent\textbf{Input:}
\begin{itemize}
    \item \textbf{Problem:} \texttt{<question>}
    \item \textbf{Reasoning Process:} \texttt{<content>}
\end{itemize}

\noindent\textbf{Output Requirements:}
\begin{enumerate}
    \item \textbf{Format:} Present the summary as a numbered list (e.g., 1., 2., 3.).
    \item \textbf{Content:} For each step, describe only the essential action to be taken (e.g., ``Calculate X,'' ``Verify Y''). Be concise and prescriptive.
    \item \textbf{Focus:} Omit explanations, justifications, or intermediate conclusions.
\end{enumerate}

\noindent\textbf{Goal:} Create a high-level plan that is easy to follow and execute.
\end{quote}

\subparagraph*{Execution Phase Prompt}
The following prompt is used to generate the detailed execution steps based on the exploration plan:

\begin{quote}
\noindent\textbf{Role:} You are a meticulous problem solver.\\
\textbf{Task:} Solve the given question by strictly following the provided guideline, showing all detailed reasoning.\\

\noindent\textbf{Input:}
\begin{itemize}
    \item \textbf{Question:} \texttt{<question>}
    \item \textbf{Guideline:} \texttt{<content>}
\end{itemize}

\noindent\textbf{Output Requirements:}
\begin{enumerate}
    \item Follow the guideline exactly, numbering each step accordingly (e.g., 1., 2., ...).
    \item Do not include any content outside the solution steps.
    \item Begin from Step 1, expanding each step with necessary calculations and logical reasoning.
    \item Conclude by placing the final answer within a `\verb|\boxed{}|` environment.
\end{enumerate}

\noindent\textbf{Important:} Ensure every mathematical or logical operation is explicitly shown.
\end{quote}

\paragraph{EF-SFT dataset Construction prompt}
The following prompt is used to extract the exploration part for EF-SFT dataset in medical domain.
\begin{quote}
\noindent\textbf{Role:} You are a professional doctor.\\
\textbf{Task:} Summarize the diagnostic reasoning process into a concise, actionable guideline.\\

\noindent\textbf{Input:}
\begin{itemize}
    \item \textbf{Question:} \texttt{<question>}
    \item \textbf{Reasoning Process:} \texttt{<content>}
\end{itemize}

\noindent\textbf{Output Requirements:}
\begin{enumerate}
    \item \textbf{Structure:} Present the summary as a numbered list (1., 2., ...), starting directly with the first step.
    \item \textbf{Conciseness:} Use no more than 5 steps. Each step must be under 15 words and state only the critical objective (e.g., “Assess cardiac function”).
    \item \textbf{Focus:} Highlight the most critical diagnostic step. Omit all explanations, justifications, or unrelated content.
\end{enumerate}

\noindent\textbf{Goal:} Create a concise and accurate diagnostic plan focused on key actions.
\end{quote}

\paragraph{LLM-Combination Prompt}

To enable the model to select the most promising exploration plan, we use the following prompt. The model is instructed to act as an impartial judge, evaluating the provided plans based on their clarity, correctness, and likelihood of leading to a successful solution.

\begin{quote}
\noindent\textbf{Role:} You are an expert mathematical reasoner and an impartial judge. Your task is to evaluate several proposed plans for solving a given math problem and identify the single best one.

\noindent\textbf{Input:}
\begin{itemize}
    \item \textbf{Problem:} \texttt{<problem>}
    \item \textbf{Candidate Plans:} A numbered list of K exploration plans.
    \texttt{
    Plan 1: $<exploration_1>$
    Plan 2: $<exploration_2>$
    ...
    Plan K: $<exploration_K>$
    }
\end{itemize}

\noindent\textbf{Instructions:}
\begin{enumerate}
    \item Carefully analyze the problem and each of the K candidate plans.
    \item Assess the plans based on their logical soundness, potential for success, and efficiency.
    \item Select the single best plan that is most likely to lead to a correct and complete solution.
\end{enumerate}

\noindent\textbf{Output Format:}
Output only the full text of the single best plan you have selected. Do not add any extra commentary, explanation, or formatting.
\end{quote}

\paragraph{Adherence Judge Prompt}
\label{llm-judge}
The following prompt is used to evaluate whether an execution strictly adheres to the provided exploration plan.

\begin{quote}
\noindent\textbf{Role:} You are a rigorous evaluator. Your task is to judge if the execution strictly follows the exploration plan.

\noindent\textbf{Input:}
\begin{itemize}
    \item \textbf{Question:} $<question>$
    \item \textbf{Exploration (PLAN):} $<exploration>$
    \item \textbf{Execution:} $<execution>$
\end{itemize}

\noindent\textbf{Evaluation Criteria:}
Provide a final score of 0, 0.5, or 1.0 based on the following:
\begin{itemize}
    \item \textbf{1.0 (Strictly Adheres):} The execution follows the exploration's logic and key steps exactly. All critical reasoning or calculations in the exploration are present and correctly implemented in the execution.
    \item \textbf{0.5 (Partially Adheres):} The execution addresses the main goal but deviates in specifics: it may skip non-critical steps, change the order of minor steps, or contain minor logical gaps while reaching a correct conclusion.
    \item \textbf{0.0 (Does Not Adhere):} The execution ignores major exploration steps, follows a different approach, contains significant logical errors, or fails to implement the exploration's core intent.
\end{itemize}

\noindent\textbf{Output Format:}
First, provide a concise rationale comparing the exploration and execution. Then, output the score strictly as:\\
\texttt{[RATIONALE]: <your analysis>}\\
\texttt{[SCORE]: <0.0, 0.5, or 1.0>}
\end{quote}
% Reasoning Example
\subsection{E\textsuperscript{2}C Reasoning Examples}
\paragraph{Example 1(Math, from AMC23): Meeting Problem}
\subparagraph*{Input Question}
Cities $A$ and $B$ are $45$ miles apart. Alicia lives in $A$ and Beth lives in $B$. 
Alicia bikes towards $B$ at 18 miles per hour. Leaving at the same time, Beth bikes 
toward $A$ at 12 miles per hour. How many miles from City $A$ will they be when they meet?
Provide the final answer in the boxed{}.

\subparagraph*{Exploration Phase}
1. Identify the speeds and initial distance between the two cities.
2. Determine the combined speed at which the distance between Alicia and Beth is closing.
3. Calculate the time it takes for them to meet by dividing the initial distance by their combined speed.
4. Use the time to calculate the distance Alicia travels from City A.

\subparagraph*{Execution Phase}
\begin{enumerate}
\item \textbf{Identify the speeds and initial distance between the two cities.}
Distance between City A and City B: 45 miles. Alicia's speed: 18 miles per hour (towards B). Beth's speed: 12 miles per hour (towards A)
\item \textbf{Determine the combined speed at which the distance between Alicia and Beth is closing.}
    Since they are moving towards each other, their speeds add up:
    \[
    \resizebox{0.90\columnwidth}{!}{$\displaystyle
    \text{Combined speed} = 18 + 12 = 30 \text{ miles per hour}$}
    \]

\item \textbf{Calculate the time it takes for them to meet by dividing the initial distance by their combined speed.}
    The time until they meet is:
    \[
    \resizebox{0.88\columnwidth}{!}{$\displaystyle
    \text{Time} = \frac{\text{Distance}}{\text{Combined speed}} = \frac{45}{30} = 1.5 \text{ hours}$}
    \]

\item \textbf{Use the time to calculate the distance Alicia travels from City A.}
    Alicia's distance from City A when they meet is:
    \[
    \resizebox{0.88\columnwidth}{!}{$\displaystyle
    \text{Distance} = \text{Speed} \times \text{Time} = 18 \times 1.5 = 27 \text{ miles}$}
    \]
\end{enumerate}

\subparagraph*{Final Answer}
$$\boxed{27}$$

\paragraph{Example 2 (Math, from AIME'24): Table Tennis Tournament Problem}

\subparagraph*{Input Question}
In a table tennis tournament every participant played every other participant exactly once. Although there were twice as many right-handed players as left-handed players, the number of games won by left-handed players was $40\%$ more than the number of games won by right-handed players. (There were no ties and no ambidextrous players.) What is the total number of games played? Provide the final answer in the boxed{}.
\subparagraph*{Exploration Phase}
1. Define variables for the number of left-handed and right-handed players.
2. Express the total number of games played in terms of the total number of players.
3. Determine the total number of games won by left-handed and right-handed players.
4. Set up an equation based on the given relationship between wins.
5. Solve the equation to find the total number of players.
6. Calculate the total number of games played.

\subparagraph*{Execution Phase}
\begin{enumerate}
\item \textbf{Define variables for the number of left-handed and right-handed players.}
Let $L$ be the number of left-handed players, and $R$ be the number of right-handed players.
Given: $R = 2L$

\item \textbf{Express the total number of games played in terms of the total number of players.}
Total players: $L + R = 3L$
Total games: $\binom{3L}{2} = \frac{3L(3L-1)}{2}$

\item \textbf{Determine the total number of games won by left-handed and right-handed players.}
Let $W_L$ be wins by left-handed players, $W_R$ be wins by right-handed players.
Total wins: $W_L + W_R = \frac{3L(3L-1)}{2}$
Given: $W_L = 1.4W_R$

\item \textbf{Set up an equation based on the given relationship between wins.}
Substitute $W_L = 1.4W_R$ into total wins:
$1.4W_R + W_R = \frac{3L(3L-1)}{2}$
$2.4W_R = \frac{3L(3L-1)}{2}$

\item \textbf{Solve the equation to find the total number of players.}
Try $L = 3$:
$W_R = \frac{3 \times 3 \times (9-1)}{4.8} = \frac{72}{4.8} = 15$
$W_L = 1.4 \times 15 = 21$
Total wins: $15 + 21 = 36$

\item \textbf{Calculate the total number of games played using the total number of players.}
Total games: $\frac{3 \times 3 \times (9-1)}{2} = \frac{72}{2} = 36$
\end{enumerate}
\subparagraph*{Final Answer}
$$\boxed{36}$$

\paragraph{Example 3 (Medical,From MMLU-Anatomy): Urinary System Question}

\subparagraph*{Input Question}
Which of the following best describes the structure that collects urine in the body?
A. Bladder
B. Kidney  
C. Ureter
D. Urethra
Please reasoning step-by-step. Provide the final answer in the boxed{}.

\subparagraph*{Exploration Phase}
1. Identify the function of each organ listed in the options.
2. Determine which organ is responsible for storing urine.
3. Eliminate options that do not fit the function of collecting urine.
4. Select the organ that best matches the description of collecting urine.
\subparagraph*{Execution Phase}
\begin{enumerate}
\item \textbf{Identify the function of each organ listed in the options:}
\begin{itemize}
\item \textbf{A. Bladder:} Hollow, muscular organ that stores urine until excretion
\item \textbf{B. Kidney:} Filters blood and produces urine, but does not store urine
\item \textbf{C. Ureter:} Transports urine from kidneys to bladder, does not store urine
\item \textbf{D. Urethra:} Carries urine from bladder to outside body, does not store urine
\end{itemize}

\item \textbf{Determine which organ is responsible for storing urine:}
The bladder is the primary organ responsible for urine storage. It expands as it fills with urine and contracts during urination.

\item \textbf{Eliminate options that do not fit the function of collecting urine:}
\begin{itemize}
\item Kidney (B): Produces urine but does not store it
\item Ureter (C): Transports urine but does not store it  
\item Urethra (D): Excretes urine but does not store it
\item Bladder (A): Only organ designed for urine collection and storage
\end{itemize}

\item \textbf{Select the organ that best matches the description of collecting urine:}
The bladder is the structure that collects and stores urine in the body.
\end{enumerate}
\subparagraph*{Final Answer}
$$\boxed{A}$$
% TODO example
\paragraph{Example 4 (Test-Time Scaling, from AIME'24): Table Tennis Tournament Problem}

This example demonstrates how E\textsuperscript{2}C test-time scaling operates over compact exploration plans rather than full reasoning chains, using the same input question from Example 2.

\subparagraph*{Input Question}
In a table tennis tournament every participant played every other participant exactly once. Although there were twice as many right-handed players as left-handed players, the number of games won by left-handed players was $40\%$ more than the number of games won by right-handed players. (There were no ties and no ambidextrous players.) What is the total number of games played? Provide the final answer in the boxed{}.

\subparagraph*{Step 1: Sample K Exploration Plans (K=4)}
First, the E\textsuperscript{2}C model is prompted to generate K=4 diverse, high-level exploration plans for the problem.

\begin{itemize}[leftmargin=15pt,topsep=3pt, itemsep=3pt]
    \item \textbf{Plan 1}
    1. Let L be the number of left-handed players, the total games are C(3L, 2). 2. Let W\_L be wins by lefties and W\_R be wins by righties.
    3. Solve this system for L, then find the total games.

    \item \textbf{Plan 2}
    1. The total number of players must be a multiple of 3, let's call it 3L. The total games played is C(3L, 2).
    2. Let's test small integer values for L (L=1, 2, 3...) and check if the resulting total games can be split into wins for left- and right-handed players satisfying the 40\% more condition.
    
    \item \textbf{Plan 3}
    1. Assume the number of wins is proportional to the number of players. Let right-handed players have W\_R wins.
    2. Left-handed players have half the number of players, so they should have W\_L wins.
    3. Set up W\_L = 1.4 * W\_R and solve based on the total number of games.
    
    \item \textbf{Plan 4}
    {1. Let the number of wins by right-handed players be W\_R. Then the wins by left-handed players is 1.4 * W\_R.
    2. The total number of games is 2.4 * W\_R. The total number of games is also given by C(3L, 2).
    3. Set C(3L, 2) = 2.4 * W\_R and find an integer solution for L.
    }
\end{itemize}

\subparagraph*{Method A: E\textsuperscript{2}C-Select (Self LM-Judge)}
The four plans above, along with the original question, are fed into the model with the Self LM-Judge prompt. The model evaluates the plans and selects the most robust and direct strategy.

\begin{enumerate}
    \item \textbf{Selection:} The Self LM-Judge identifies Plan 1 as the most comprehensive and logically sound approach, as it correctly sets up the system of equations from first principles.
    \item \textbf{Execution:} A single execution is performed, conditioned only on Plan 1. This execution proceeds exactly as detailed in Example 2, arriving at the correct answer.
\end{enumerate}
\textbf{Final Answer (Self LM-Judge)}: $\boxed{36}$

\subparagraph*{Method B: E\textsuperscript{2}C-ReAct Loop}
This feedback-driven method executes the selected plan, and only returns to the exploration stage if execution fails to produce a valid final answer.

\begin{enumerate}
    \item \textbf{Initial Selection:} The Self LM-Judge again selects Plan 1 as the initial exploration because it directly sets up the algebraic relationship between left- and right-handed wins.
    \item \textbf{Execution Attempt:} The model executes the selected plan deterministically. If the execution reaches a valid boxed answer, the loop stops immediately.
    \item \textbf{Exploration Refinement on Failure:} If execution stalls, loops, or fails to produce a boxed answer, the recent execution trace is summarized as feedback. The model then revises the exploration plan rather than generating a new full solution. For example, it may add the missing constraint that the total number of games equals both $\binom{3L}{2}$ and $W_L+W_R$.
    \item \textbf{Fallback to Unused Plans:} If the revised plan is too similar to the failed one, the method switches to an unused candidate plan that avoids the stuck point.
    \item \textbf{Retry Execution:} The revised or replacement exploration is executed again. In this example, the corrected plan yields the same final result:
    \begin{itemize}
        \item Let $W_L = 1.4 W_R$ and $W_L+W_R = \binom{3L}{2}$.
        \item Enforce integrality of the number of players and wins.
        \item Solve the resulting constraints to obtain the total number of games.
    \end{itemize}
\end{enumerate}
\textbf{Final Answer (E\textsuperscript{2}C-ReAct Loop)}: $\boxed{36}$

% Extra Experiment
\subsection{Pure Prompt-based E\textsuperscript{2}C}
\label{subsec:exploration-execution}

\begin{table}[b]
\centering
\caption{Pass@5 accuracy (\%) for different numbers of sampled explorations \(K\).}
\label{tab:pass5_K}
\small
\begin{tabular}{lcccc}
\toprule
Dataset & \(K=2\) & \(K=3\) & \(K=4\) & \(K=5\) \\
\midrule
MATH500  & \(\mathrm{84.4}\) & \(\mathrm{83.2}\) & \(\mathrm{84.0}\) & \(\mathrm{84.0}\) \\
AIME24   & \(\mathrm{26.7}\) & \(\mathrm{33.0}\) & \(\mathrm{36.7}\) & \(\mathrm{26.7}\) \\
AIME25   & \(\mathrm{23.3}\) & \(\mathrm{30.0}\) & \(\mathrm{30.0}\) & \(\mathrm{26.7}\) \\
\bottomrule
\end{tabular}
\end{table}
We product an experiment with pure prompt-based E\textsuperscript{2}C on Qwen3-8B. For each problem we first sample \(K\) independent \emph{exploration} traces by prompting the model \(K\) times with a short exploration prompt; each exploration is a concise (2--4 short sentence) reasoning sketch that does not contain the final answer. We then combine the \(K\) explorations into a single execution prompt (providing the problem and the numbered explorations) and ask the model to produce one final \emph{Execution:} section that computes the final answer. Performance is reported as pass@5 for different values of \(K\).The results are much worse than the E\textsuperscript{2}C model with E\textsuperscript{2}C-(SFT+RL), which demonstates that a prompt engeneering is not enough.

\paragraph{Exploration prompt}
\label{prompt:exploration}
The following prompt was used to generate each individual exploration (one exploration per model call).

\begin{quote}
\noindent\textbf{Role:} You are a careful math problem solver.

\noindent\textbf{Input:}
\begin{itemize}
    \item \textbf{Problem:} \texttt{<problem>}
\end{itemize}

\noindent\textbf{Instructions:}
\begin{itemize}
    \item Produce exactly one short reasoning sketch (an \emph{exploration}) that helps approach the problem.
    \item The exploration must be concise (about 2–4 short sentences).
    \item Do \textbf{not} produce the final answer in this call.
    \item Stop immediately after the single exploration text and do not append any extra commentary, labels, or formatting.
\end{itemize}

\noindent\textbf{Output format:} A single short exploration paragraph (2–4 short sentences) and nothing else.
\end{quote}

\paragraph{Execution prompt}
\label{prompt:execution}
The following prompt was used to synthesize the \(K\) independently sampled explorations into a final execution.

\begin{quote}
\noindent\textbf{Role:} You are a careful math problem solver.

\noindent\textbf{Input:}
\begin{itemize}
    \item \textbf{Problem:} \texttt{<problem>}
    \item \textbf{Explorations:} \\
    Exploration 1: \texttt{<exploration 1>} \\
    Exploration 2: \texttt{<exploration 2>} \\
    \(\vdots\) \\
    Exploration \{\(K\)\}: \texttt{<exploration K>}
\end{itemize}

\noindent\textbf{Instructions:}
\begin{itemize}
    \item Learn from the provided \(\{K\}\) numbered explorations and combine their useful reasoning to compute the final answer.
    \item Produce a single \textbf{Execution:} section that carries out the computation and presents the final answer.
    \item Stop immediately after the final answer. Do not append extra commentary, explanations, or any additional text beyond the required Execution section and the answer.
\end{itemize}
\end{quote}

\section*{Errata (v2 $\to$ v3)}
\label{app:errata}

An earlier version of this paper (arXiv v2) reported incorrect test-time scaling results due to an experimental error (a bug in the evaluation script).
Specifically, the following results in Table~3 were affected:

\begin{itemize}
  \item The Greedy CoT accuracy was erroneously reported as 40.6\% (corrected: 36.7\%).
  \item The E\textsuperscript{2}C-Select (Self LM-Judge) accuracy at $K{=}32$ was erroneously reported as 58.1\% (corrected: 40.0\%).
  \item Baseline results (Self-Consistency, Tree-of-Thoughts) were similarly affected.
\end{itemize}

In addition, the set of methods in the test-time scaling comparison was revised: Two new E\textsuperscript{2}C variants(E\textsuperscript{2}C-ToT and E\textsuperscript{2}C-ReAct Loop) were introduced to provide a more systematic comparison against their full-chain counterparts. All results in this version have been re-verified.

\end{document}